\documentclass[10pt,twocolumn,letterpaper]{article}

\usepackage{iccv}
\usepackage{times}
\usepackage{epsfig}
\usepackage{graphicx}
\usepackage{amsmath}
\usepackage{amssymb}
\usepackage{comment}
\usepackage{colortbl}
\usepackage{makecell}
\usepackage{pifont}
\usepackage{amssymb}
\usepackage{booktabs}
\usepackage{graphicx}
\usepackage{rotating}
\usepackage{multirow}
\usepackage[export]{adjustbox}

\usepackage[breaklinks=true,bookmarks=false]{hyperref}

\iccvfinalcopy 

\def\iccvPaperID{****} 

\ificcvfinal\fi

\begin{document}

\title{H-Net: Unsupervised Attention-based Stereo Depth Estimation Leveraging Epipolar Geometry}

\author{Baoru Huang$^{1}$\thanks{Baoru Huang and Jian-Qing Zheng contribute equally to this paper}\\%
\and Jian-Qing Zheng$^{2*}$\thanks{{\tt\small jianqing.zheng@kennedy.ox.ac.uk}}%
\and Stamatia Giannarou$^{1}$ \and Daniel S. Elson$^{1}$\\
$^{1}$The Hamlyn Centre for Robotic Surgery, Imperial College London, U.K.\\
$^{2}$The Kennedy Institute of Rheumatology, University of Oxford, U.K.
%
}

\maketitle
\ificcvfinal\thispagestyle{empty}\fi


\begin{abstract}
Depth estimation from a stereo image pair has become one of the most explored applications in computer vision, with most of the previous methods relying on fully supervised learning settings. However, due to the difficulty in acquiring  accurate and scalable ground truth data, the training of fully supervised methods is challenging. As an alternative, self-supervised methods are becoming more popular to mitigate this challenge. In this paper, we introduce the H-Net, a deep-learning framework for unsupervised stereo depth estimation that leverages epipolar geometry to refine stereo matching. For the first time, a Siamese autoencoder architecture is used for depth estimation which allows mutual information between the rectified stereo images to be extracted. To enforce the epipolar constraint, the mutual epipolar attention mechanism has been designed which gives more emphasis to correspondences of features which lie on the same epipolar line while learning mutual information between the input stereo pair. Stereo correspondences are further enhanced by incorporating semantic information to the proposed attention mechanism. More specifically, the optimal transport algorithm is used to suppress attention and eliminate outliers in areas not visible in both cameras.  Extensive experiments on KITTI2015 and Cityscapes show that our method outperforms the state-of-the-art unsupervised stereo depth estimation methods while closing the gap with the fully supervised approaches.

\end{abstract}

\section{Introduction}
Humans are remarkably capable of inferring the 3D structure of a real world scene even over short timescales. For example, when navigating along a street, we are able to locate obstacles and vehicles in motion and avoid them with a fast response time. Years of substantial interest in geometric computer vision has not accomplished comparable modeling capabilities to humans for various real-world scenes where reflections, occlusions, non-rigidity and textureless areas exist. So what can human ability be attributed to? A central concept is that humans learn the regularities of the world while interacting with it, moving around, and observing vast quantities of scenes. Consequently, we develop a rich, consistent and structural understanding of the world, which is utilized when we perceive a new scene. Our binocular vision is a supporting feature, from which the brain can not only build disparity maps, but can also combine to obtain structural information. These two ideas prompt one of the fundamental problems in computer vision --- depth estimation --- whose quality has a direct influence on various application scenarios, such as autonomous driving, robotic navigation, augmented reality and 3D reconstruction. 

Thanks to advanced deep learning techniques, the performance of depth estimation methods has improved significantly over the last few years. Most previous work relies on ground-truth depth data and considers deep architectures for generating depth maps in a supervised manner \cite{eigen2014depth, kendall2017end, luo2016efficient, xu2020aanet}. However, collecting vast and varied training datasets with accurate per-pixel ground truth depth data for supervised learning is a formidable challenge. To overcome this limitation, some recent works have shown that self-supervised methods are instead able to effectively tackle the depth estimation task \cite{Godard_2017_CVPR} \cite{pilzer2018unsupervised}. We are particularly inspired by the approaches proposed in \cite{godard2019digging, zhou2017unsupervised_ego, garg2016unsupervised, johnston2020self} where they took view synthesis as supervisory signals to train the network and exploited differences between the original input, synthesized view, and left and right disparity results as penalties (\textit{i.e.} a photometric image reconstruction cost, a left-right consistency cost and a disparity smoothness cost) to force the system to generate accurate disparity maps. However, although some works have tried to emphasize the complementary information in the stereo image pair and used sharing weights when extracted features from input images \cite{pilzer2018unsupervised} \cite{chang2018pyramid}, the contextual information between the multiple views --- especially some strong feature matches --- lie on the epipolar line, and this information has not been effectively explored and exploited. 

In this paper, we follow the unsupervised learning setting and introduce the H-Net, a novel end-to-end trainable network for depth estimation given rectified stereo image pairs. The proposed H-Net effectively fuses the information in the stereo pair and combines epipolar geometry with learning-based depth estimation approaches. In summary, our main contributions in this paper are:
\begin{itemize}
\item We design a Siamese encoder-Siamese decoder network architecture, which fuses the complementary information in the stereo image pairs while enhancing the communication between them. To the best of our knowledge, this is the first time this architecture is used for depth estimation.

\item We propose a mutual epipolar attention module to enforce the epipolar constraint in feature matching and emphasise the strong relationship between the features located along the same epipolar lines in rectified stereo image pairs. 

\item We further enhance the proposed attention module by using the optimal transport algorithm to incorporate in a novel fashion semantic information and filter out outlier feature correspondences.

\end{itemize}
We demonstrate the effectiveness of our approach on the challenging KITTI \cite{geiger2012we} and Cityscapes datasets \cite{cordts2016cityscapes}. Compared to previous approaches, the H-Net achieves state-of-the-art results.

\section{Related work}
Estimating depth maps from stereo images has been explored for decades \cite{barnard1982computational}. Accurate stereo depth estimation plays a critical role in perceiving the 3D geometric configuration of scenes and facilitating a variety of computer vision applications in the real world \cite{joung2019unsupervised}. Recent work has shown that depth estimation from a stereo image pair can be effectively tackled by learning-based methods with convolutional neural networks (CNNs) \cite{chang2018pyramid}.

\subsection{Supervised Depth Estimation}

A pyramid stereo matching network was proposed in \cite{chang2018pyramid}, where spatial pyramid pooling and dilated convolution were adopted to enlarge the receptive fields, while a stacked hourglass CNN was designed to further boost the utilization of global context information. Duggal \etal \cite{duggal2019deeppruner} proposed a differentiable PatchMatch module to abandon most disparities without requiring full cost volume evaluation, and thus the specific range to prune for each pixel could be learned. Kusupati \etal \cite{kusupati2020normal} improved the depth quality by leveraging the predicted normal maps and a normal estimation model, and proposed a new consistency loss to refine the depths from depth/normal pairs.

The above methods are fully supervised and rely on having large amounts of accurate ground truth depth for training. However, this is challenging to obtain data in various real-world settings \cite{zhou2017unsupervised_stereo_matching}. Synthetic training data is a potential solution  \cite{gaidon2016virtual} \cite{mayer2018makes}, but still requires manual curation for every new application scenario.

\subsection{Unsupervised Depth Estimation}
Due to the lack of per-pixel ground truth depth data, self-supervised depth estimation is an alternative, where image reconstruction is the supervisory signal during training \cite{godard2019digging}. The input for this type of model is usually a set of images, either as stereo pairs \cite{pilzer2018unsupervised}\cite{pilzer2019progressive}  or as monocular sequences \cite{zhou2017unsupervised_ego} \cite{johnston2020self}. 

Gard \etal \cite{garg2016unsupervised} proposed an approach using a calibrated stereo camera pair setup for unsupervised monocular depth estimation, in which depth was generated as an intermediate output and the supervision signal came from the reconstruction combining the counterpart image in a stereo pair. Godard \etal \cite{Godard_2017_CVPR} extended this work by using forward and backward reconstructions of different image views while adding an appearance matching loss and multi-scale loss to the model. Per-pixel minimum reprojection loss and auto-masking were explored in \cite{godard2019digging}, which allowed the network to ignore objects moving at the same velocity as the camera or frames captured when the camera was static, with further improved results. Johnston \etal \cite{johnston2020self} introduced discrete disparity prediction and applied self-attention to the depth estimation framework, providing a more robust and sharper depth estimation map. 

It has been shown that training with an added binocular color image could help single image depth estimation by posing it as an image reconstruction problem without requiring ground truth \cite{Godard_2017_CVPR} \cite{godard2019digging}. Andrea \etal \cite{pilzer2018unsupervised} showed that the depth estimation results could be effectively improved within an adversarial learning framework, with a deep generative network that learned to predict the disparity map for a calibrated stereo camera using a wrapping operation. 

In the multi-view (stereo) depth estimation task, it is naturally to employ complementary features from different views to establish the geometric correspondences.
Zhou \etal \cite{zhou2017unsupervised_stereo_matching} presented a framework that learned stereo matching costs without human supervision, in which the network parameters were updated in an iterative manner and a left-right check was applied to guide the training procedure. Joung \etal\cite{joung2019unsupervised} proposed a framework to compute matching cost in an unsupervised setting, where the putative positive samples in every training iteration were selected by exploiting the correspondence consistency between two stereo images.
Although these methods tried to explore the feature relationship between the stereo images, the concrete matching matrix have not been effectively exploited and been applied to the learning procedure, which leads to a cost of details and a waste of geometric information, especially the strong constraints on the epipolar line.
\begin{figure}[t]
\begin{center}

\includegraphics[width=\linewidth]{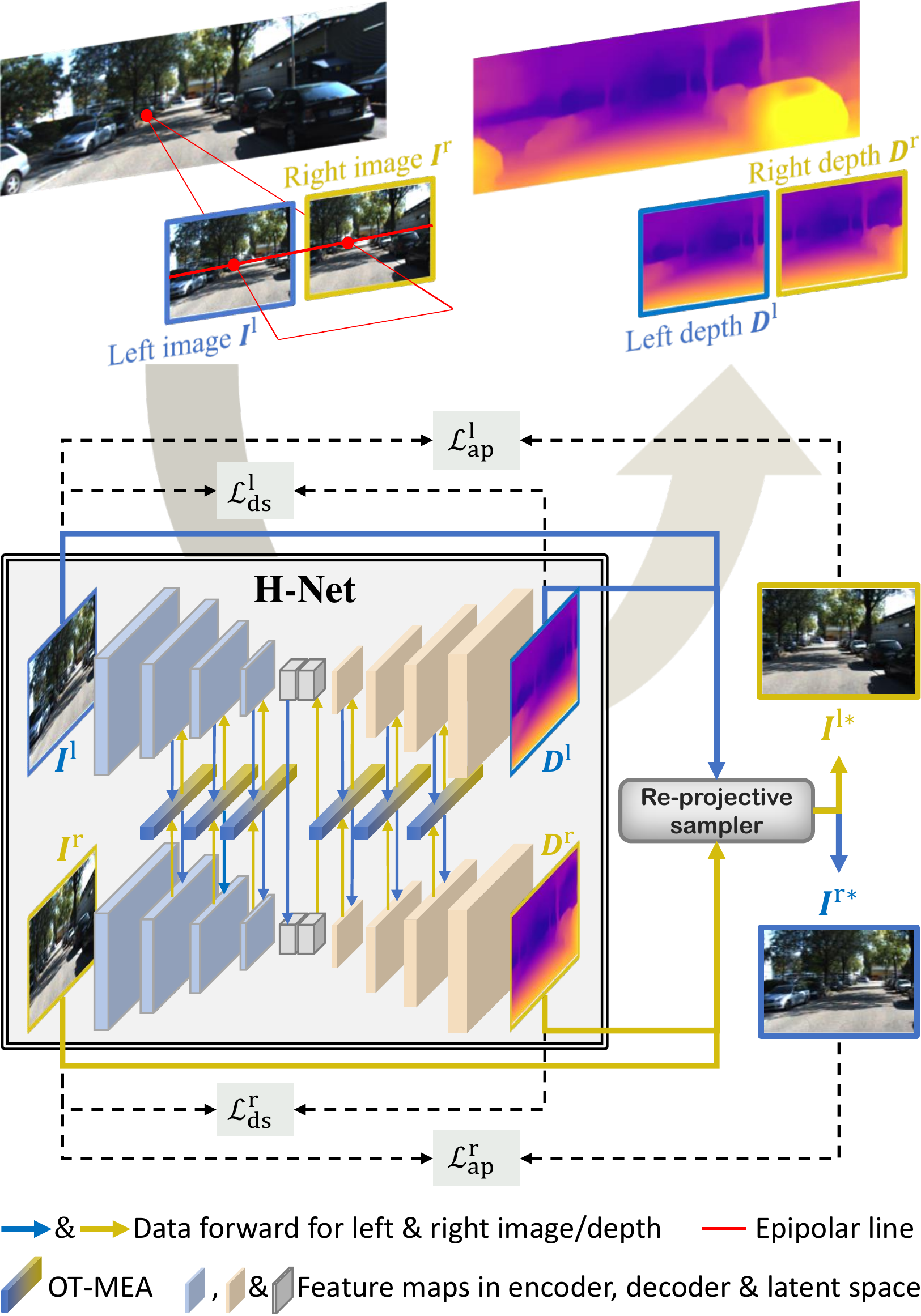}
\end{center}
\caption{The H-Net architecture.}
  
\label{fig:pipeline}
\end{figure}

\begin{figure*}
\begin{center}

\includegraphics[width=\linewidth]{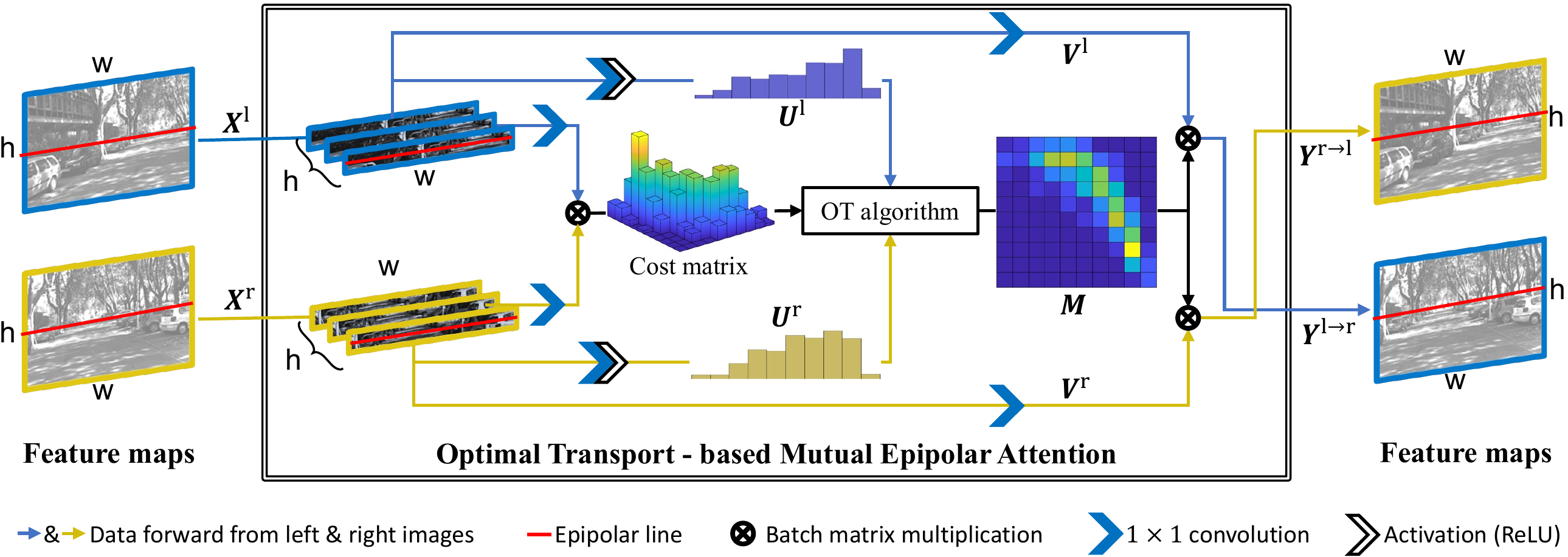}
\end{center}
   \caption{Optimal Transport based Mutual Epipolar Attention (OT-MEA) block combines OT retrieving (Eq.~\ref{eq:ot-phi}) into the MEA module (Eq.~\ref{eq:ma}).}
\label{fig:ot-mea}
\end{figure*}

\section{Method}
Here, we describe the details of the process of depth prediction using the proposed H-Net.

\subsection{H-Net architecture}
\label{sec:hnet}
In this paper, the encoder-decoder structure Monodepth2 \cite{godard2019digging} was adopted as the fundamental backbone, based on the U-Net \cite{ronneberger2015u}.
As shown in Fig.~\ref{fig:pipeline}, the proposed architecture consisted of a double-branch encoder and a double-branch decoder. 
To make the network compact, similar to \cite{chang2018pyramid} and \cite{pilzer2018unsupervised}, a Siamese Encoder - Siamese Decoder (SE-SD) structure was designed with shared weights between the two branches in both the encoder and the decoder. To our knowledge, this is the first time a SE-SD is used for stereo depth estimation enabling the extraction of mutual information from the pair of input images.

The Siamese Encoder (SE) of H-net included two branches of Resnet18 \cite{he2016deep} with shared trainable parameters. 
The left and right rectified images $\textit{\textbf{I}}^{\rm l},\textit{\textbf{I}}^{\rm r}\in\mathbb{R}^{3\times h_0\times w_0}$ were fed into each branch of the SE to extract common features from the input images, where $h_0,w_0$ denotes the image size. The outputs of the three deeper Residual-down-sampling (Res-down) blocks in the SE were interconnected with a novel mutual attention block proposed in this work --- the so-called Optimal Transport-based Mutual Epipolar Attention (OT-MEA) block, shown in Fig.~\ref{fig:ot-mea} and explained in detail below. 

The abstract latent features from the encoder were fused in the middle part by concatenating the feature maps extracted from each SE block between the two branches. 
Each concatenated map is then convolved by two separate convolution layers with different trainable parameters.

The decoder took the fused latent features as inputs and generated sigmoid outputs for each input image similar to \cite{Godard_2017_CVPR} and \cite{godard2019digging}. It was composed of the same number of Residual-up-sampling (Res-up) blocks as Res-down to recover the full resolution, as well as OT-MEA blocks inserted in the first three Res-up blocks. 
Each sigmoid output  $\Omega$ of the decoder was transformed to scene depth as $D = 1/(a\Omega +b)$. The parameters a and b were selected to constrain depth $D$ between 0.1 and 100 units.

\subsection{Mutual Epipolar Attention}
\label{sec:mea}

State-of-the-art deep learning methods for stereo depth estimation have not considered the epipolar constraint when estimating feature correspondences. In this work, we introduce a mutual attention mechanism to give more emphasis to features correspondences which lie on the same epipolar line.

Recently, Wang et al. \cite{wang2018non} proposed the Non-Local (NL) block which allowed them to exploit global-range attention in an image sequence. This was then extended with the introduction of the Mutual NL (MNL) block \cite{zheng2020d} to explore the mutual relationships between different inputs in multi-view vision. However, global-range feature matching in the NL and MNL blocks suffers from the high number of parameters, memory requirement and training time. Furthermore, these blocks can be misled by repeated textures in the scenes.

To overcome the above limitations, we designed the Mutual Epipolar Attention (MEA) module to constrain feature correspondences to the same epipolar line between a pair of rectified stereo images. Here, MEA was defined as:
\begin{equation}
\label{eq:ma}
\left\{
\begin{array}{cc}
\textbf{\textit{Y}}^{\rm l\to r}:=\Psi({\textbf{\textit{X}}^{\rm l}})\otimes\Phi({\textbf{\textit{X}}^{\rm l}},{\textbf{\textit{X}}^{\rm r}})\\
\textbf{\textit{Y}}^{\rm r\to l}:=\Psi({\textbf{\textit{X}}^{\rm r}})\otimes\Phi({\textbf{\textit{X}}^{\rm r}},{\textbf{\textit{X}}^{\rm l}})
\end{array}
\right.
\end{equation}
where $\otimes$ denotes the batch matrix multiplication, $\textbf{\textit{X}}^{\rm l},\textbf{\textit{X}}^{\rm r}\in\mathbb{R}^{h \times c\times w}$ denote the transported and reshaped input signals from the two branches, $\textbf{\textit{Y}}^{\rm l\to r},\textbf{\textit{Y}}^{\rm r\to l}\in\mathbb{R}^{h \times c\times w}$ are the output signals from the MEA block. $\Phi:\mathbb{R}^{h \times c\times w}\times\mathbb{R}^{h \times c\times w}\to\mathbb{R}^{h\times w\times w},(\textbf{\textit{X}}^{\rm 1},\textbf{\textit{X}}^{\rm 2})\mapsto \textbf{\textit{M}}^{1\to 2}$ is a pair-wise matching function, the so called retrieval function, which evaluates the compatibility between the two inputs. $\Psi:\mathbb{R}^{h \times c\times w}\to\mathbb{R}^{h \times c\times w},\textbf{\textit{X}}\mapsto \textbf{\textit{V}}$ is a unary function which maps vectors from one feature space to another which is essential for fusion. 

Following the settings in \cite{wang2018non}, the Embedded Gaussian (EG) similarity representation was used to define our matching function:
\begin{equation}
\label{eq:ma-eg}
\Phi_{\rm EG}(\textbf{\textit{X}}^1,\textbf{\textit{X}}^2):={\rm softmax}(\mathcal{C}_1({{\textbf{\textit{X}}^1})^{\top}\otimes\mathcal{C}_2(\textbf{\textit{X}}^2}))  
\end{equation}
where $\mathcal{C}$ is the $1\times 1$ convolution,
and was also used in the unary function for vector mapping:
\begin{equation}
\Psi:=\mathcal{C}
\end{equation}

In our experimental work, the EG-based MEA and MNL modules were compared and denoted as EG-MEA and EG-MNL, respectively.

\subsection{Optimal transport based mutual attention}
\label{sec:otma}

In stereo vision, the input images have been captured from cameras at different positions and view angles making the field of view of the two images sightly different. This can cause outliers in depth estimation due to incorrect feature correspondences in the areas which are not visible to both cameras. To eliminate outliers in these areas, we further enhanced our proposed MEA module to suppress the contribution of correspondences in these occluded areas during feature matching. The EG similarity representation defined in Eq.(\ref{eq:ma-eg}) cannot achieve this because all the areas of the input signals are equally considered. 

For this purpose, we formulated the matching task in Eq.(\ref{eq:ma}) as an optimal transport (OT) problem as it has already been proven that OT improves semantic correspondence \cite{liu2020semantic}. Thus, a new OT-based retrieval function is further proposed, tailored to our stereo depth estimation problem:
\begin{equation}
\begin{array}{l}
\Phi_{\rm OT}(\textbf{\textit{X}}^1,\textbf{\textit{X}}^2):=\mathop{\arg\min}\limits_{\textbf{\textit{M}}}{\|\textbf{\textit{M}}\odot{\rm e}^{1-\mathcal{C}'_1({{\textbf{\textit{X}}^1})^{\top}\otimes\mathcal{C}'_2(\textbf{\textit{X}}^2})}\|}_1
\\
{\rm s.t.}~~~~~~~~~~~\textbf{u}\otimes\textbf{\textit{M}}=\Theta(\textbf{\textit{X}}^2),~\textbf{u}\otimes\textbf{\textit{M}}^\top
=\Theta(\textbf{\textit{X}}^1)
\end{array}
\label{eq:ot-phi}
\end{equation}
where $\odot$ denotes a Hadamard product, $\mathcal{C}'$ is a sequence operation of convolution and channel-wise Euclidean normalization, $\textbf{u}\in\{1\}^{h\times 1\times w}$ is a matrix with all elements equal to 1. $\Theta:\mathbb{R}^{h \times c\times w}\to\mathbb{R}^{h \times 1\times w},\textit{\textbf{X}}\mapsto\textbf{\textit{U}}$ is the sequence operation of convolution, ReLU activation and pixel-wise L1-normalization to generate the transported mass of pixels $\textbf{\textit{U}}$. The matrix $\textbf{\textit{M}}$ is the variable to be optimised and represents the optimal matching matrix $\textbf{\textit{M}}^{1\to 2}$.

Here, OT-based matching in Eq.~(\ref{eq:ot-phi}) assigns to each pixel the sum of each column of the similarity weights in matching matrix $\textbf{\textit{M}}^{1\to 2}$, which is constrained by the mass:
\begin{equation}
\left\{
\begin{array}{l}
U_{ij}^{1}=\sum_{k}{M_{ijk}^{1\to 2}}\\
U_{ik}^{2}=\sum_{j}{M_{ijk}^{1\to 2}}
\end{array}
\right.
,\forall i,j,k\in\mathbb{Z},i\leq\textit{h},j,k\leq\textit{w}
\label{eq:ot-mass}
\end{equation}
where $U_{ij}^{1}$, $U_{ik}^{2}$ and ${M_{ijk}^{1\to 2}}$ are the elements of the $\textbf{\textit{U}}^{1}$, $\textbf{\textit{U}}^{2}$ and $\textbf{\textit{M}}^{1\to 2}$ respectively indexed by $i,j,k$.
In contrast to the equal consideration by EG-based matching in Eq.~(\ref{eq:ma-eg}), varying weights are assigned to different correspondences in Eq.~(\ref{eq:ot-mass}), determined by the latent semantic messages forwarded from the input signals. This enables the OT module to suppress the outliers and focus on correspondences with more mass which lie on the semantic areas.

In this paper, since Eq.~(\ref{eq:ot-phi}) is a convex optimization problem, the Sinkhorn algorithm is used to obtain the numerical solution of this OT problem \cite{liu2020semantic}.
OT matching based MEA is denoted as OT-MEA and Fig.~\ref{fig:ot-mea} illustrates the implementation sketch of the OT-MEA used in H-net. Both MEA and OT modules can be used separately or in combination and we present their impact with an ablation study in Section~\ref{ablation_section}. OT-MEA was also compared in our experimental work to the OT matching based MNL, denoted as OT-MNL.

\subsection{Self-Supervised Training}
For the left and right input images $\textit{\textbf{I}}^{\rm l}, \textit{\textbf{I}}^{\rm r}\in\mathbb{R}^{3\times h_0\times w_0}$, the sigmoid outputs of the H-Net were transformed to depth maps $\textit{\textbf{D}}^{\rm l},\textit{\textbf{D}}^{\rm r}\in\mathbb{R}^{1\times h_0\times w_0}$ as explained in Section~3.1. By combining one of the depth maps (e.g $\textit{\textbf{D}}^{\rm l}$) and the count-part input image ($\textit{\textbf{I}}^{\rm r}$), we were able to reconstruct the initial image ($\textit{\textbf{I}}^{\rm {l*}}$) using the re-projection sampler \cite{jaderberg2015spatial}.  Here we used the left image $\textit{\textbf{I}}^{\rm l}$ as an example to present the supervisory signal and loss components. The final loss function included the loss terms for both left and right images. The similarity between the input image $\textit{\textbf{I}}^{\rm l}$ and the reconstructed image $\textit{\textbf{I}}^{\rm {l*}}$ provides our supervisory signal.  Our photometric error function $\mathcal{L}^l_{ap}$ was defined as the combination of \textit{$L_1$}-norm and structural similarity index (SSIM) \cite{godard2019digging}:

\begin{equation} 
\label{eq:appearance_matching_loss}
\mathcal{L}_{\rm ap}^{\rm l} = \frac{1}{N}\sum_{i, j} \frac{\gamma}{2} (1-{\rm SSIM}(I_{ij}^{\rm l}, I_{ij}^{l*})) + (1-\gamma) {\|I_{ij}^{\rm l}-I_{ij}^{l*}\|}_1
\end{equation}
where, \(N\) denotes the number of pixels and \(\gamma\) is the weighting for \textit{$L_1$}-norm loss term. 
To improve the predictions around object boundaries, an edge-aware smoothness term $\mathcal{L}_{ds}$ was applied \cite{godard2019digging,johnston2020self}:

\begin{equation}
\mathcal{L}_{\rm ds}^{\rm l} = \frac{1}{N}\sum_{i,j}|\partial_x (\textit{d}_{ij}^{\rm {l*}})|e^{-|\partial_x {\textit{I}}_{ij}^{\rm l}|} + |\partial_y (\textit{d}_{ij}^{\rm {l*}})| e^{-|\partial_y {\textit{I}}_{ij}^{\rm l}|}
\end{equation}
where ${d^{l*}=d^l\sqrt{d^l}}$ represents the mean-normalized inverse of depth ($1/\textbf{\textit{D}}$) which aims at preventing shrinking of the depth prediction \cite{wang2018learning}. 

To overcome the gradient locality of the re-projection sampler, we adopted the multi-scale estimation method presented in \cite{godard2019digging}, which first upsamples the low resolution depth maps (from the intermediate layers) to the input image resolution and then reprojects and resamples them. The errors were computed at the higher input resolution. Finally, the photometric loss and per-pixel smoothness loss were balanced by the smoothness term $\lambda$ and the total loss was averaged over each scale (s), branch (left and right) and batch:
\begin{equation} \label{eq:total_depth_loss}
\begin{split}
\mathcal{L}_{\rm total} &=\frac{1}{2m}\sum_{s=1}^m (\mathcal{L}_s^{\rm l} + \mathcal{L}_s^{\rm r})\\  &=
\frac{1}{2m} \sum_{s=1}^m \big((\mathcal{L}_{\rm ap}^{\rm l} + \lambda \mathcal{L}_{\rm ds}^{\rm l}) + (\mathcal{L}_{\rm ap}^{\rm r} + \lambda \mathcal{L}_{\rm ds}^{\rm r}))
\end{split}
\end{equation}
\label{sec:training}

\section{Experiments}

\begin{table*}[ht!]
\resizebox{\textwidth}{!}{
\begin{tabular}{|c|c||c|c|c|c|c|c|c|}
\hline

\hline
Method  & Train & \cellcolor[RGB]{255,170,170}Abs Rel & \cellcolor[RGB]{255,170,170}Sq Rel & \cellcolor[RGB]{255,170,170}RMSE  & \cellcolor[RGB]{255,170,170}RMSE log & \cellcolor[RGB]{153,204,255}${\delta < 1.25 }$ &\cellcolor[RGB]{153,204,255} $\delta < 1.25^2$  &\cellcolor[RGB]{153,204,255}${\delta < 1.25^3}$ \\ 
\hline

\hline
Eigen \cite{eigen2014depth}  & D  & 0.203   & 1.548  & 6.307 & 0.282    & 0.702   & 0.890   & 0.890   \\
Liu \cite{liu2015learning} &D &0.201 &1.584 &6.471 &0.273 &0.680 &0.898 &0.967 \\

AdaDepth \cite{kundu2018adadepth} &D* &0.167 &1.257 &5.578 &0.237 &0.771 &0.922 &0.971 \\
Kuznietsov \cite{kuznietsov2017semi} &DS &0.113 &0.741& 4.621 &0.189 &0.862 &0.960 &\underline{0.986} \\
DVSO \cite{yang2018deep} &D*S &0.097 &0.734 &4.442 &0.187 &0.888 &0.958 &0.980 \\
SVSM FT \cite{luo2018single} &DS &\underline{0.094} &\underline{0.626} &4.252 &0.177 &0.891 &0.965 &0.984 \\
Guo \cite{guo2018learning} &DS &0.096 &0.641 &\underline{4.095} &\underline{0.168} &\underline{0.892} &\underline{0.967} &\underline{0.986} \\
DORN \cite{fu2018deep} &D &\textbf{0.072} &\textbf{0.307} &\textbf{2.727} &\textbf{0.120} &\textbf{0.932} &\textbf{0.984} &\textbf{0.994} \\ \hline

UnDeepVO \cite{li2018undeepvo} &MS &0.183 &1.730 &6.57 &0.268 &- &- &-  \\
Zhan FullNYU \cite{zhan2018unsupervised} &D*MS &0.135 &1.132 &5.585 &0.229 &0.820 &0.933 &0.971  \\
EPC++ \cite{luo2019every} &MS &0.128 &0.935 &5.011 &0.209 &0.831 &0.945 &\underline{0.979}  \\

Monodepth2 \cite{godard2019digging} &MS &\underline{0.106} &0.818 &4.750 &0.196 &0.874 &\underline{0.957} &\underline{0.979}  \\
Monodepth2 (1024 × 320) \cite{godard2019digging} &MS &\underline{0.106} &\underline{0.806} &\underline{4.630} &\underline{0.193} &\underline{0.876} &\textbf{0.958} &\textbf{0.980}  \\
Yang \cite{yang2020d3vo}   &MS &\textbf{0.099} &\textbf{0.763} &\textbf{4.485} &\textbf{0.185}  &\textbf{0.885} &\textbf{0.958} &\underline{0.979}\\
\hline
Garg \cite{garg2016unsupervised}† &S &0.152 &1.226 &5.849 &0.246 &0.784 &0.921 &0.967 \\
Monodepth R50 \cite{Godard_2017_CVPR}† &S &0.133 &1.142 &5.533 &0.230 &0.830 &0.936 &0.970 \\
StrAT \cite{mehta2018structured} &S &0.128 &1.019 &5.403 &0.227 &0.827 &0.935 &0.971 \\
3Net (R50) \cite{poggi2018learning} &S &0.129 &0.996 &5.281 &0.223 &0.831 &0.939 &0.974 \\
3Net (VGG) \cite{poggi2018learning} &S &0.119 &1.201 &5.888 &0.208 &0.844 &0.941 &0.978 \\
Pilzer \cite{pilzer2018unsupervised} &S &0.152  &1.388 &6.016 &0.247 &0.789 &0.918 &0.965\\
SuperDepth + pp \cite{pillai2019superdepth} (1024 × 382) &S &0.112 &0.875 &4.958 &0.207 &0.852 &0.947 &0.977 \\
Monodepth2 \cite{godard2019digging} &S &0.109 &0.873 &4.960 &0.209 &0.864 &0.948 &0.975 \\
Monodepth2 (1024 × 320) \cite{godard2019digging} &S &0.107 &0.849 &4.764 &0.201 &0.874 &0.953 &0.977 \\ 
PFN \cite{pilzer2019progressive} &S &0.102 &0.802 &4.657 &0.196 &0.882 &0.953 &0.977\\

\rowcolor[RGB]{230,230,230} \textbf{H-Net (Ours)} &S  &\underline{0.094}  &\textbf{0.600}  &   \underline{4.197}  & \underline{0.174}  & \underline{0.909}  &\underline{0.964} &\textbf{0.983} \\
\rowcolor[RGB]{230,230,230} \textbf{H-Net (Ours) Full Eigen} &S &\textbf{0.076}  &\underline{0.607}  &\textbf{4.025}  &\textbf{0.166}  &\textbf{0.918}  &\textbf{0.966}  &\underline{0.982}  \\

\hline

\hline
\end{tabular}}
\vspace{0.5mm}
\caption{Quantitative results. Comparison of our proposed H-Net to existing methods on KITTI2015 \cite{geiger2012we} using the Eigen split unless marked with `Full Eigen', which indicates the full Eigen dataset. The best result in each category are presented in bold while the second best results are \underline{underlined}. All results here are shown without post-processing \cite{Godard_2017_CVPR} unless marked with -+pp. The supervision mode for each method is indicated in the \textit{Train}: D-Depth supervision, D*-Auxiliary depth supervision, M-Self-supervised mono supervision and S-self-supervised stereo supervision. Symbol † represents the new results from github. Metrics labeled by red mean \textit{lower is better} while labeled by blue mean \textit{higher is better}.}
\label{quantitive results}
\end{table*}



\begin{figure*}[ht!]
  \centering
  \centering
  \label{Fig:final_result}
  \tiny
  \setlength{\tabcolsep}{0.18em}
  \begin{tabular}{ccccc}
      \rotatebox[origin=c]{90}{Left input} 
      &\includegraphics[width=0.235\textwidth, valign=c]{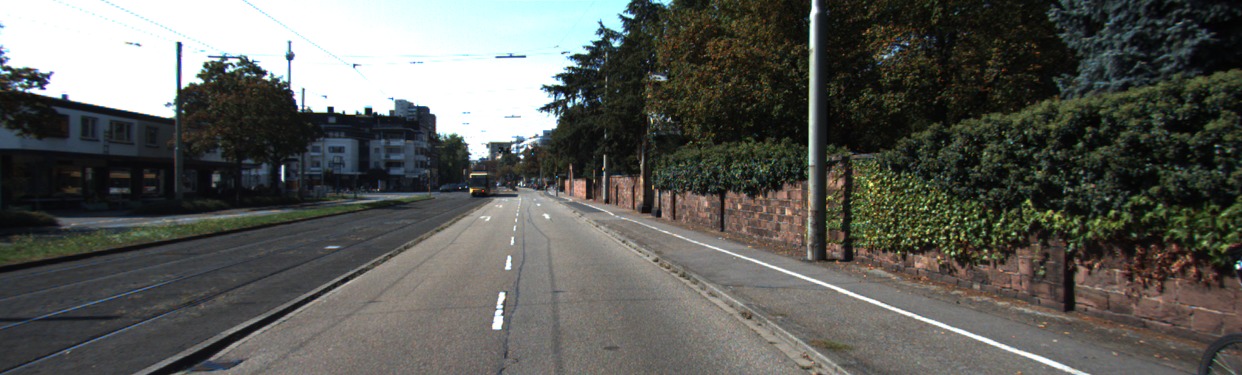}
      &\includegraphics[width=0.235\textwidth, valign=c]{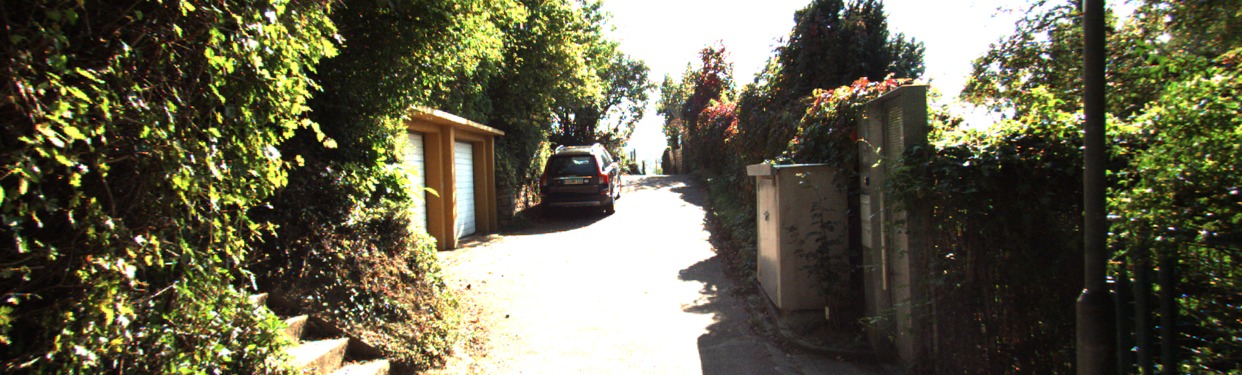}
      &\includegraphics[width=0.235\textwidth, valign=c]{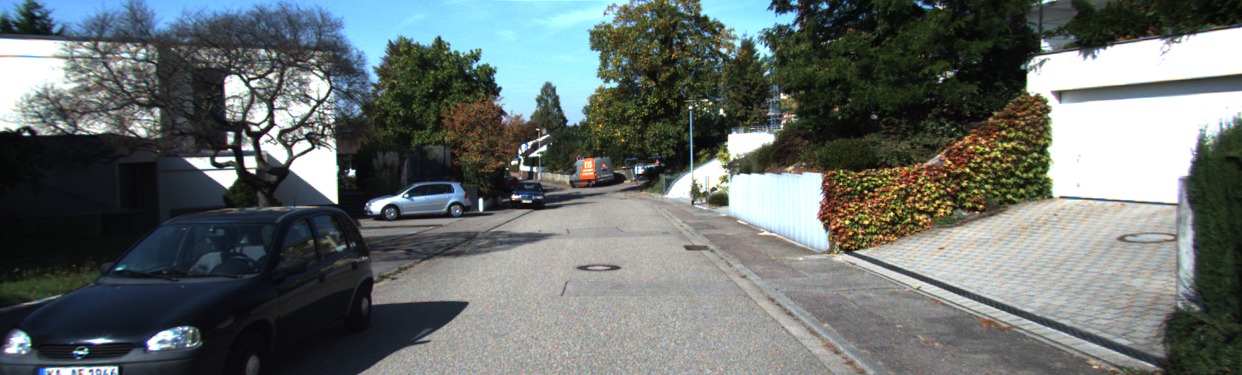}
      &\includegraphics[width=0.235\textwidth, valign=c]{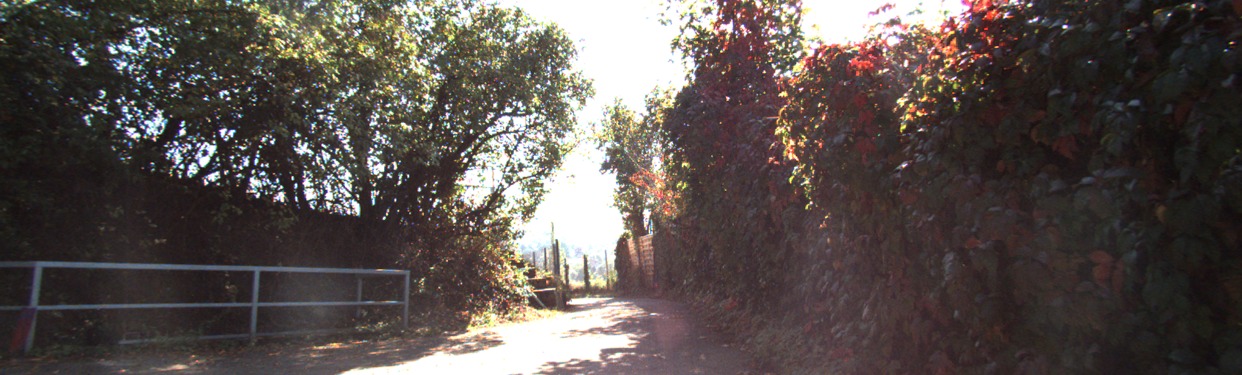}
      \vspace{0.8mm}\\
      \rotatebox[origin=c]{90}{Right input}
      &\includegraphics[width=0.235\textwidth, valign=c]{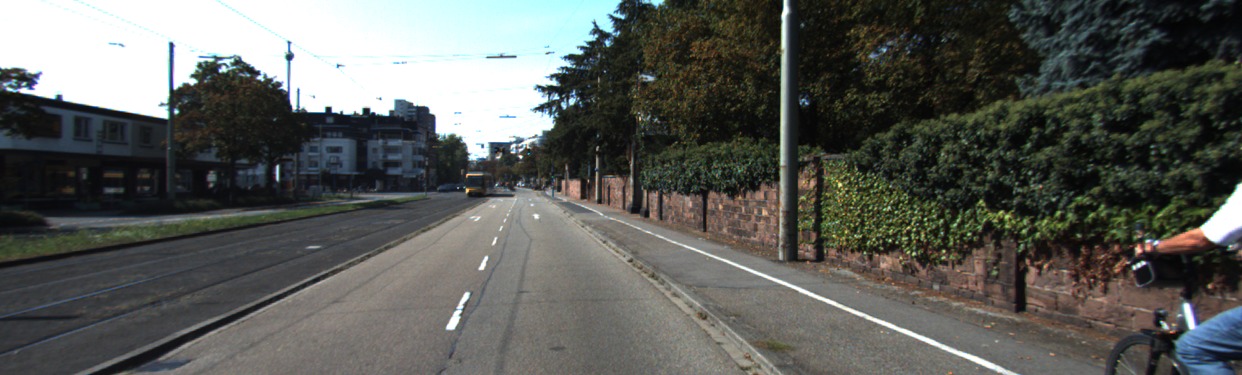}
      &\includegraphics[width=0.235\textwidth, valign=c]{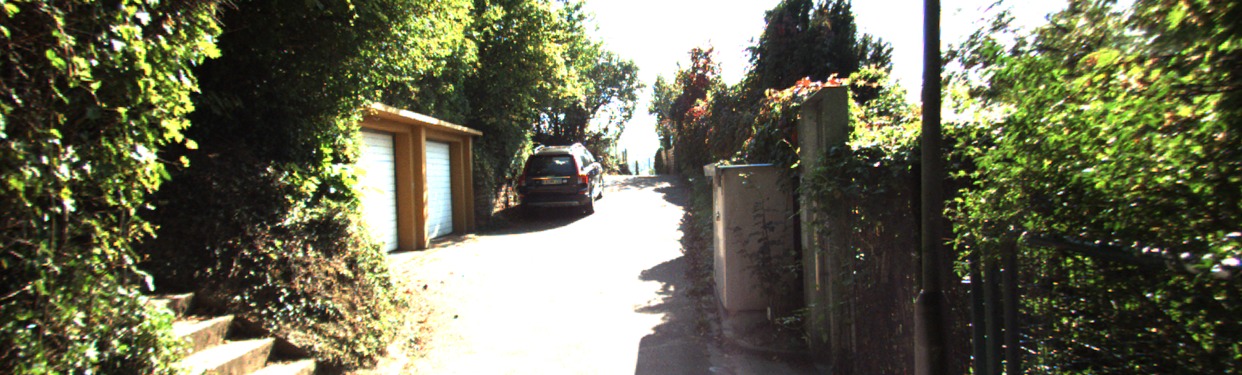}
      &\includegraphics[width=0.235\textwidth, valign=c]{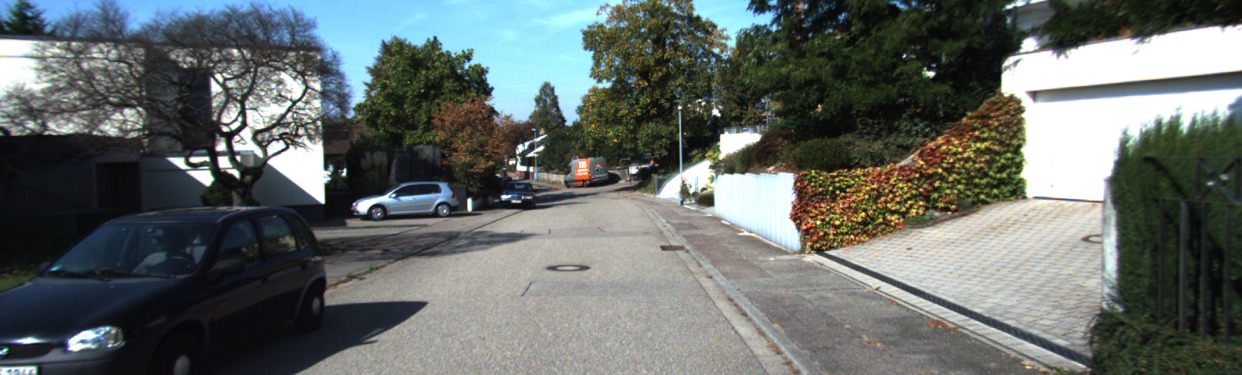}
      &\includegraphics[width=0.235\textwidth, valign=c]{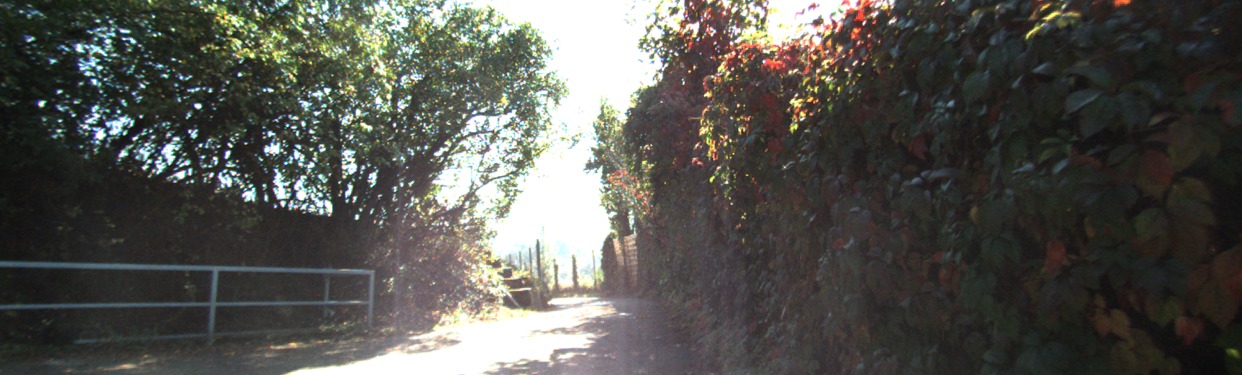}
      \vspace{0.8mm}\\
      \rotatebox[origin=c]{90}{Monodepth\cite{Godard_2017_CVPR}}
      &\includegraphics[width=0.235\textwidth, valign=c]{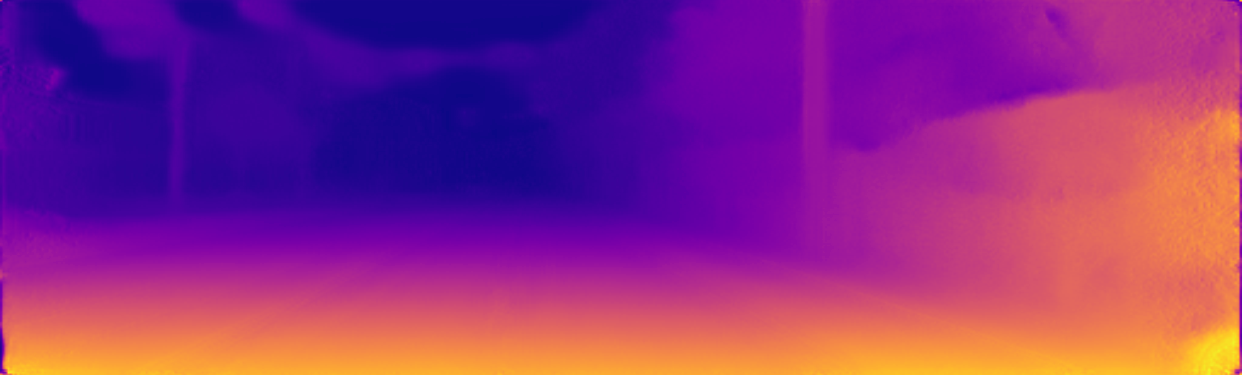}
      &\includegraphics[width=0.235\textwidth, valign=c]{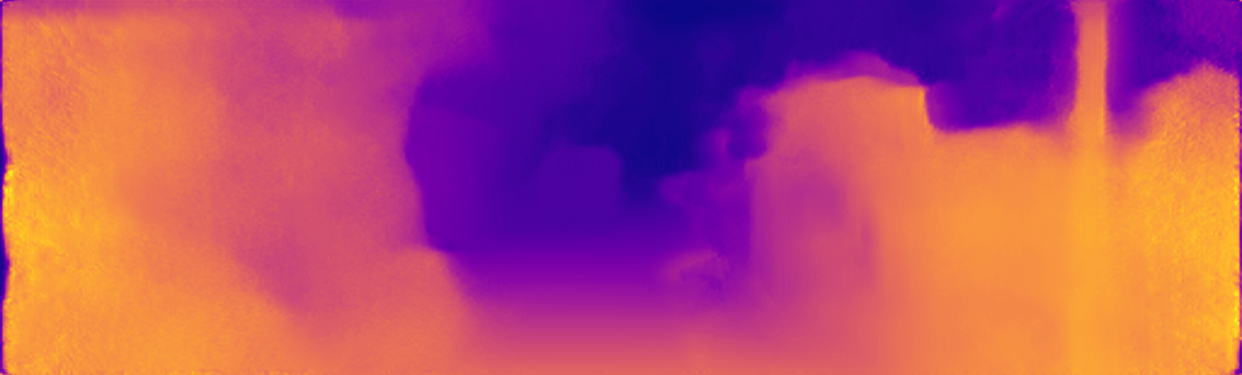}
      &\includegraphics[width=0.235\textwidth, valign=c]{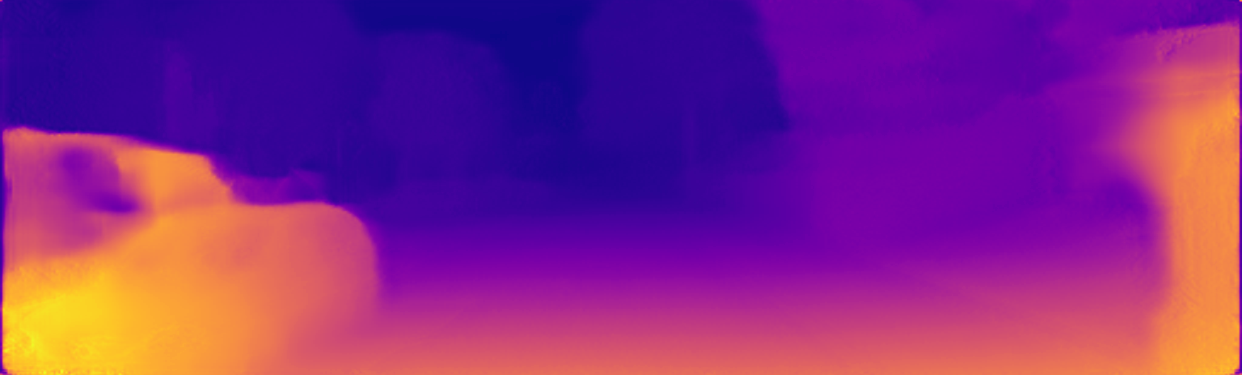}
      &\includegraphics[width=0.235\textwidth, valign=c]{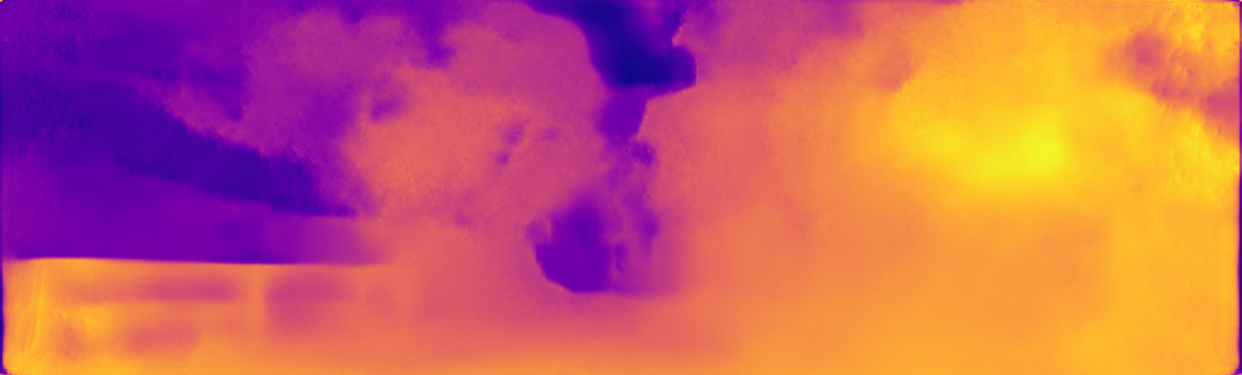}\\
      \rotatebox[origin=c]{90}{Monodepth2(S)\cite{godard2019digging}}
      &\includegraphics[width=0.235\textwidth, valign=c]{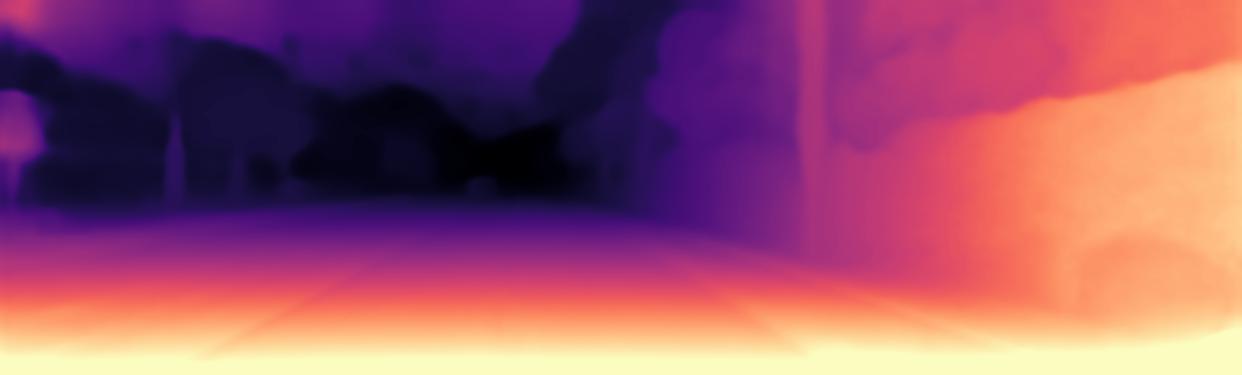}
      &\includegraphics[width=0.235\textwidth, valign=c]{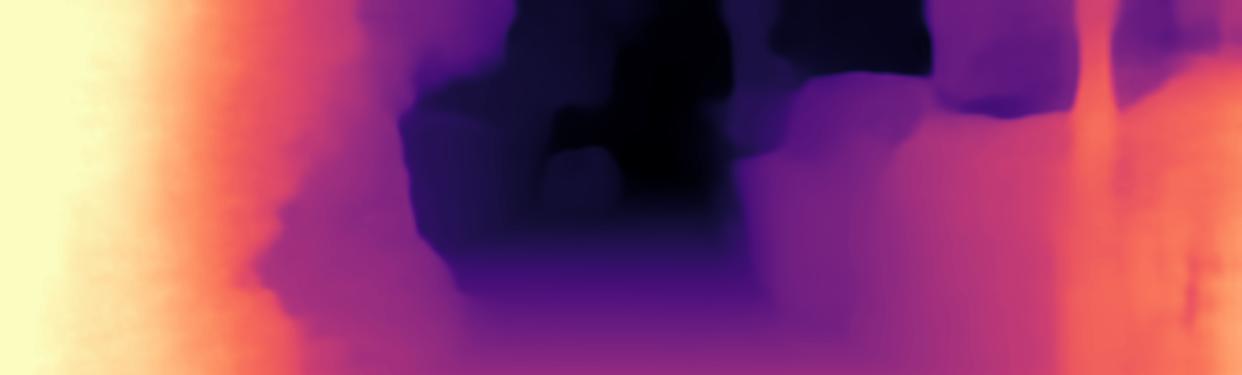}
      &\includegraphics[width=0.235\textwidth, valign=c]{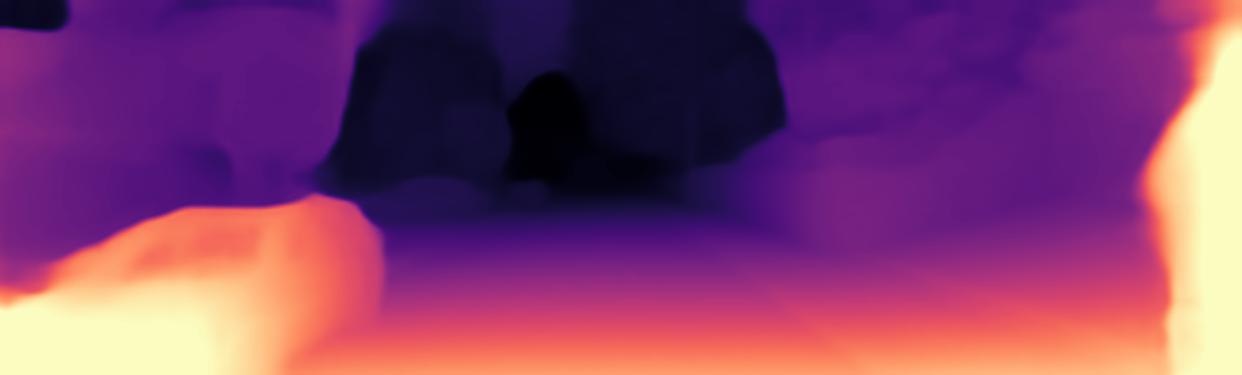}
      &\includegraphics[width=0.235\textwidth, valign=c]{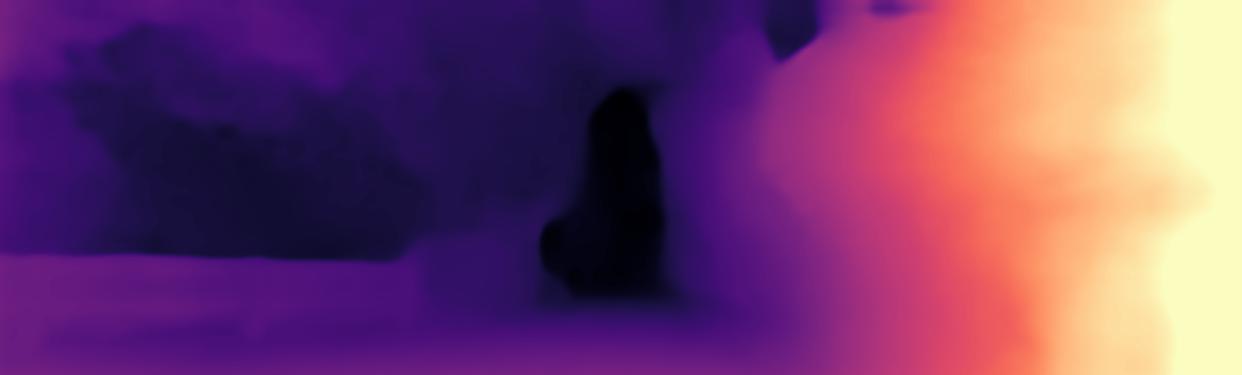}\\
      \rotatebox[origin=c]{90}{Pilzer\cite{pilzer2019progressive}}
      &\includegraphics[width=0.235\textwidth, valign=c]{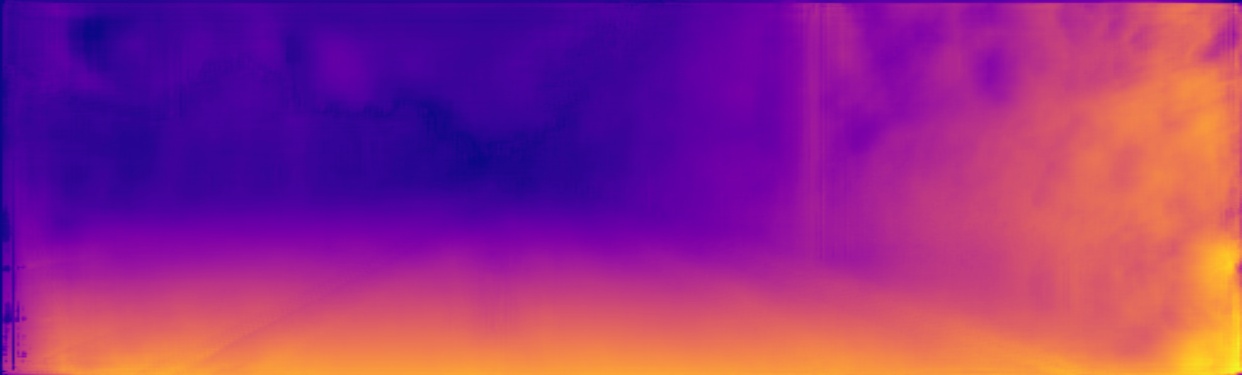}
      &\includegraphics[width=0.235\textwidth, valign=c]{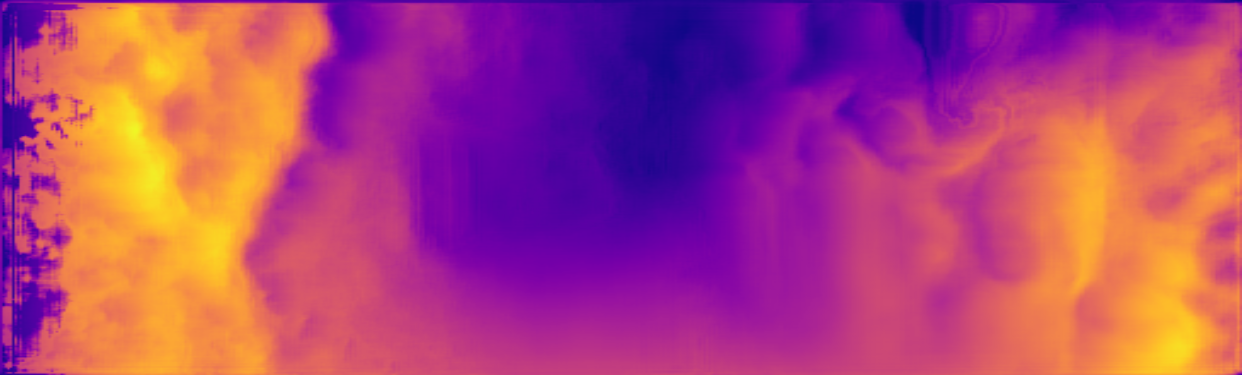}
      &\includegraphics[width=0.235\textwidth, valign=c]{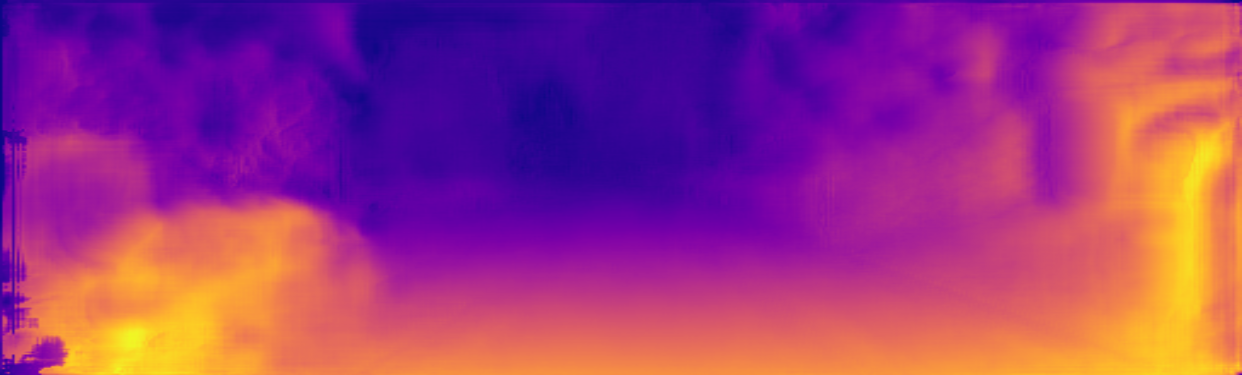}
      &\includegraphics[width=0.235\textwidth, valign=c]{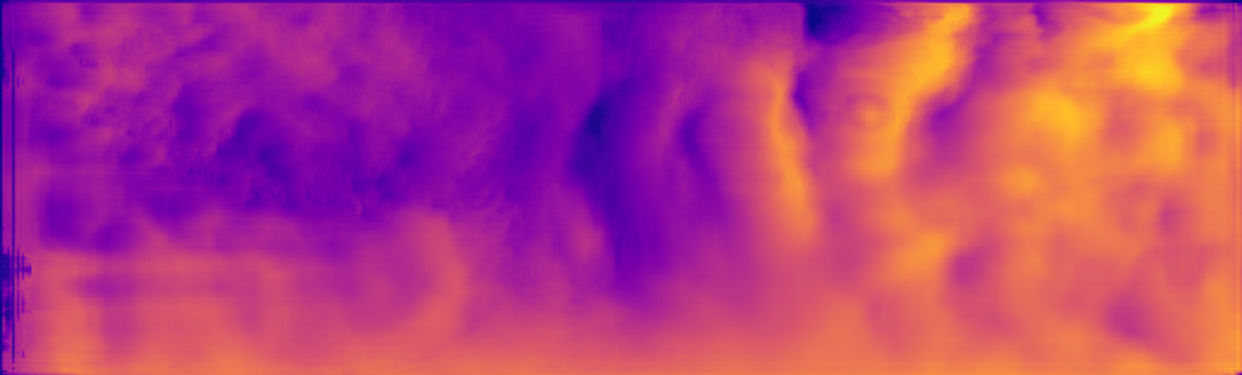}
      \vspace{1mm}\\
      \rotatebox[origin=c]{90}{\textbf{H-NET(ours)}}
      &\includegraphics[width=0.235\textwidth, valign=c]{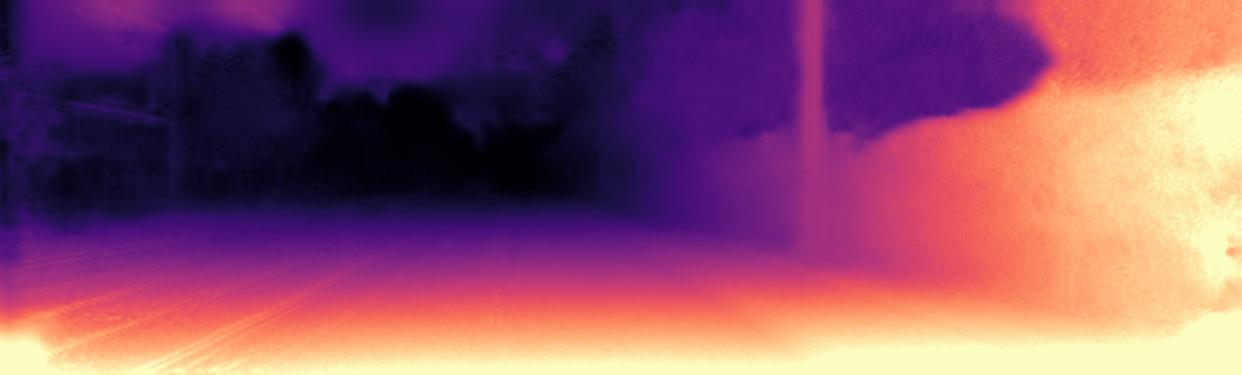}
      &\includegraphics[width=0.235\textwidth, valign=c]{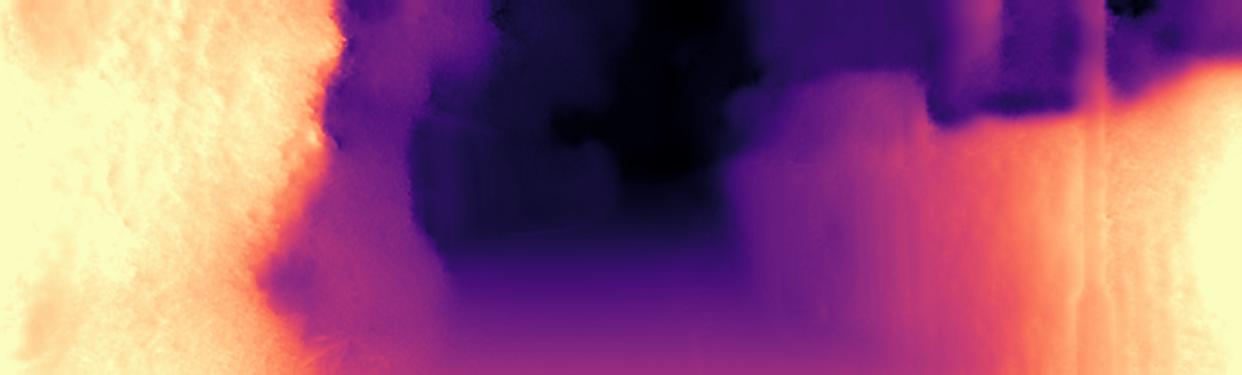}
      &\includegraphics[width=0.235\textwidth, valign=c]{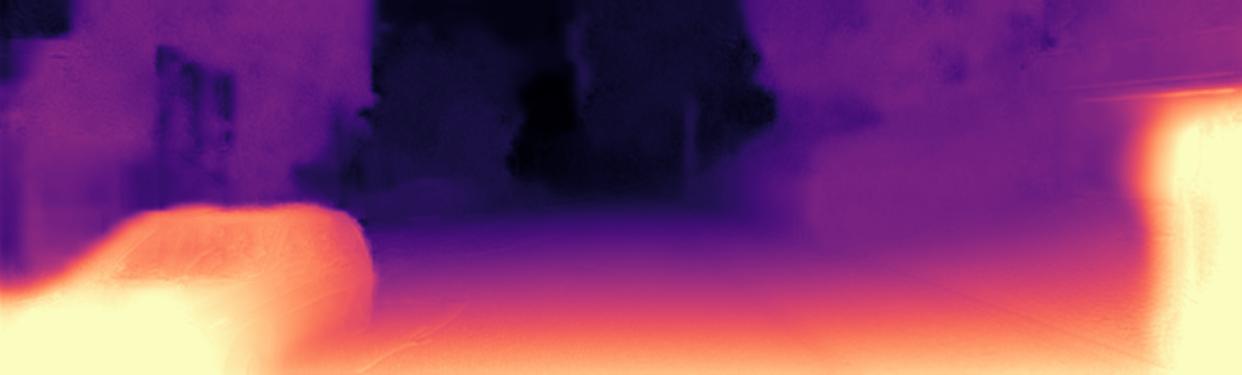}
      &\includegraphics[width=0.235\textwidth, valign=c]{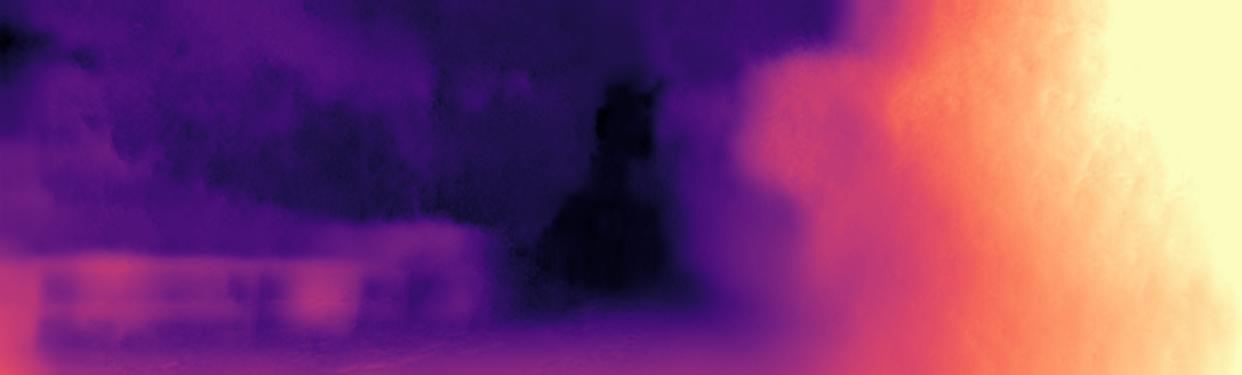}
      \vspace{1mm}\\
      
      \rotatebox[origin=c]{90}{Left input}
      &\includegraphics[width=0.235\textwidth, valign=c]{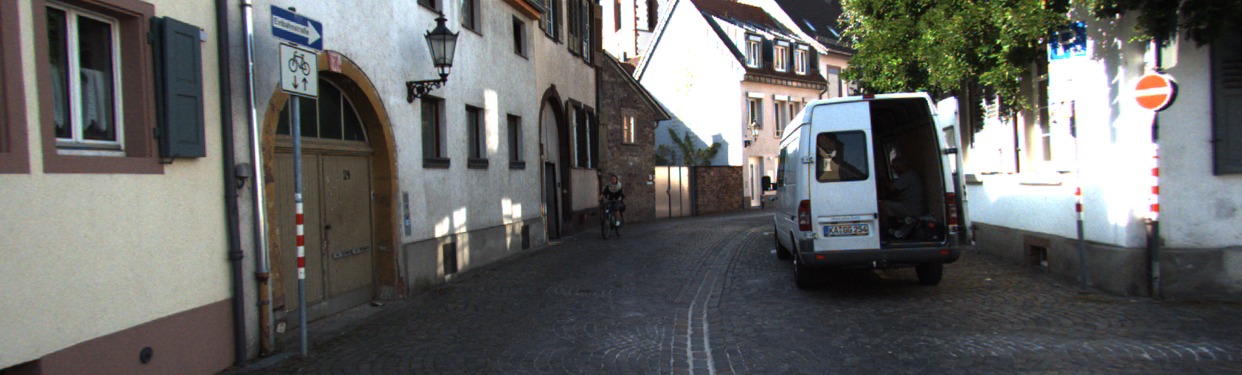}
      &\includegraphics[width=0.235\textwidth, valign=c]{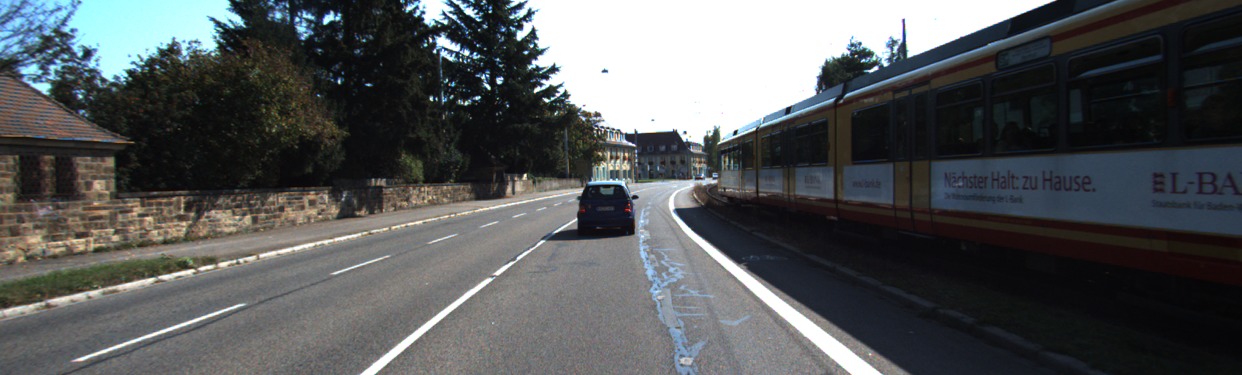}
      &\includegraphics[width=0.235\textwidth, valign=c]{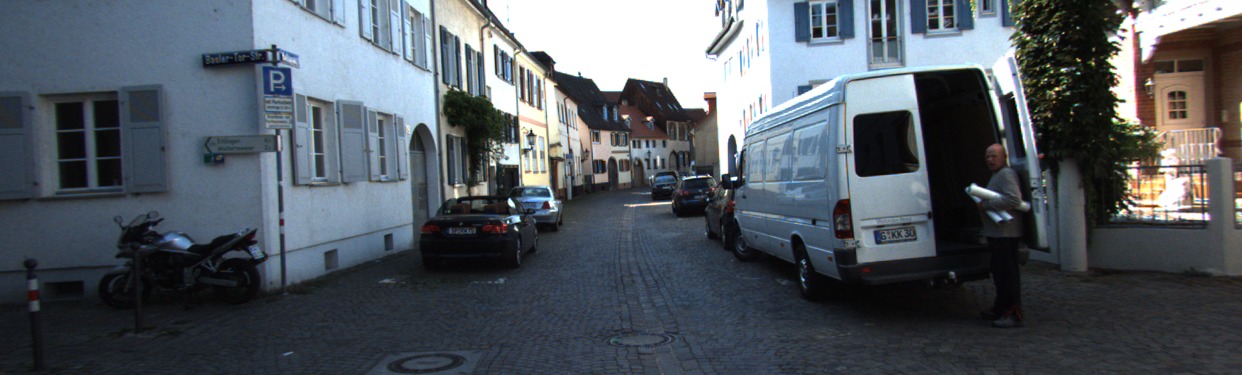}
      &\includegraphics[width=0.235\textwidth, valign=c]{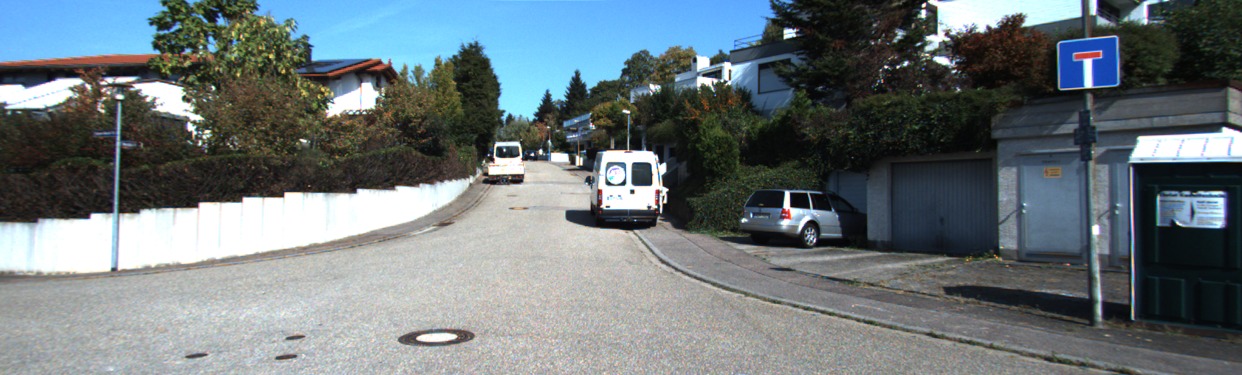}
      \vspace{0.8mm}\\
      \rotatebox[origin=c]{90}{Right input}
      &\includegraphics[width=0.235\textwidth, valign=c]{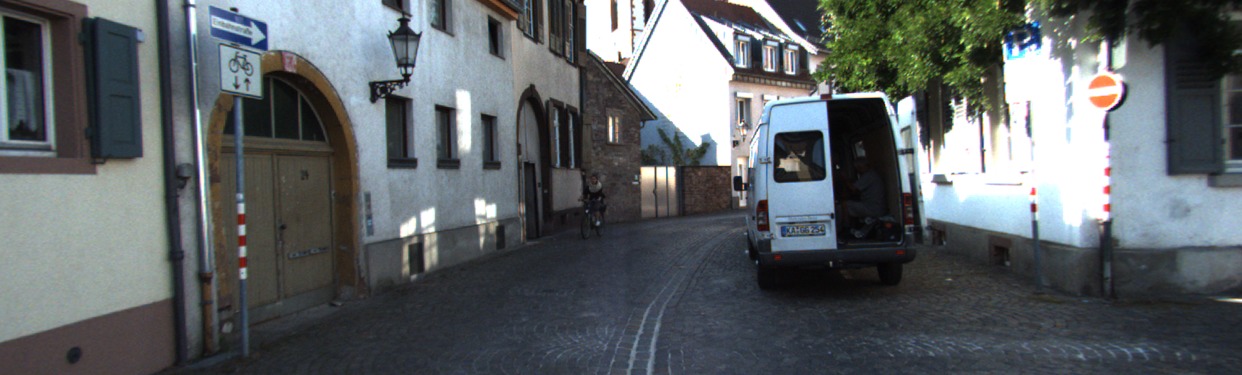}
      &\includegraphics[width=0.235\textwidth, valign=c]{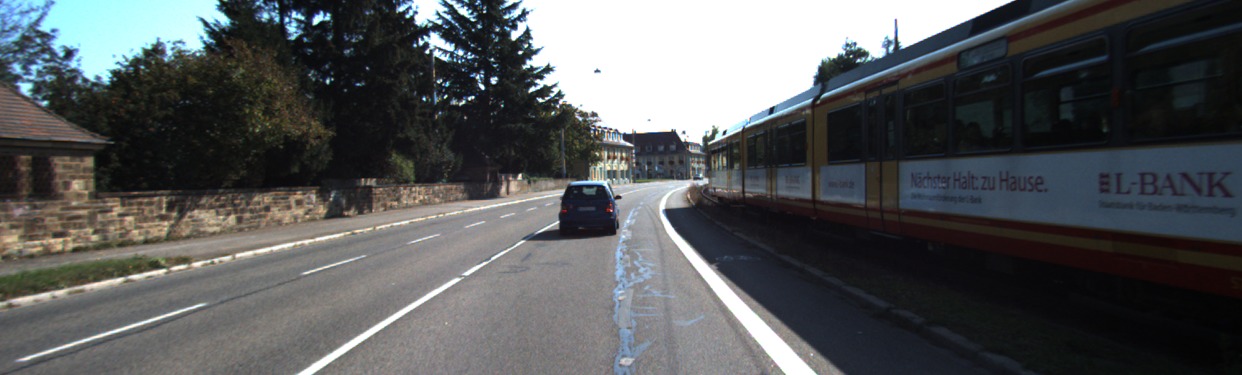}
      &\includegraphics[width=0.235\textwidth, valign=c]{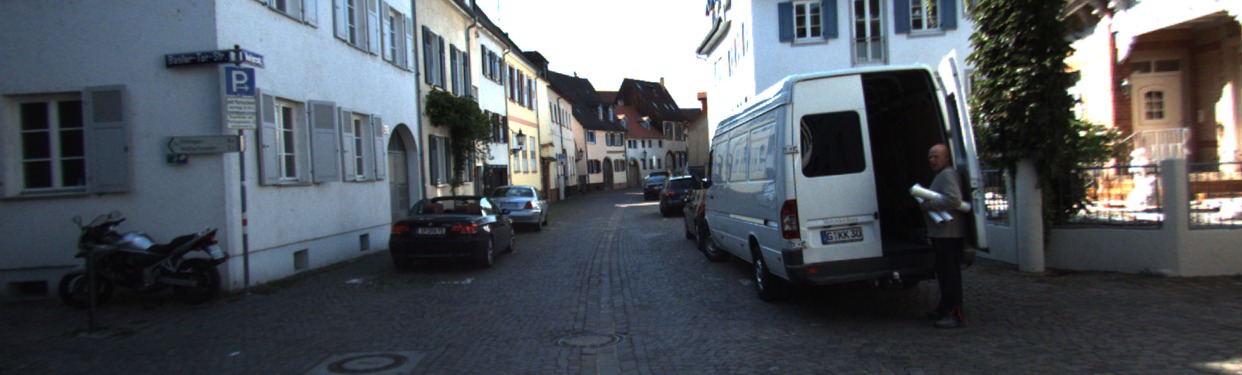}
      &\includegraphics[width=0.235\textwidth, valign=c]{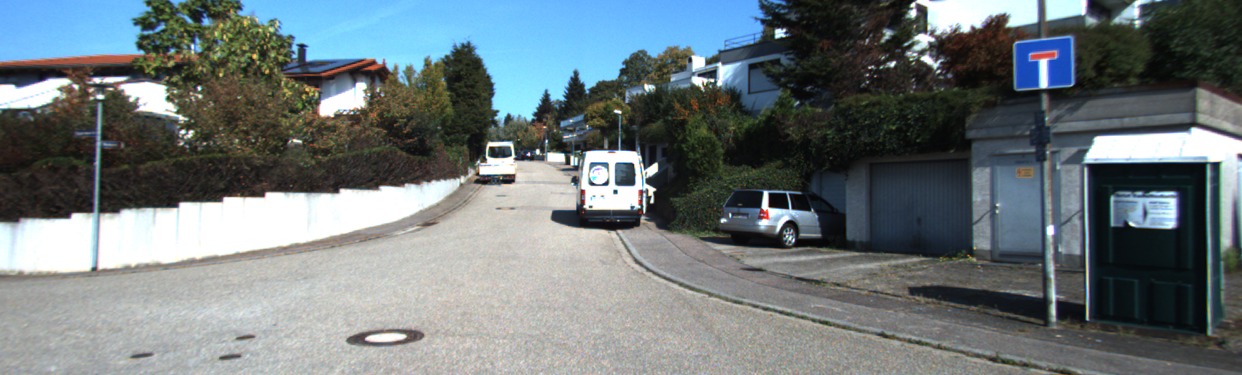}
      \vspace{0.8mm}\\
      \rotatebox[origin=c]{90}{Monodepth\cite{Godard_2017_CVPR}}
      &\includegraphics[width=0.235\textwidth, valign=c]{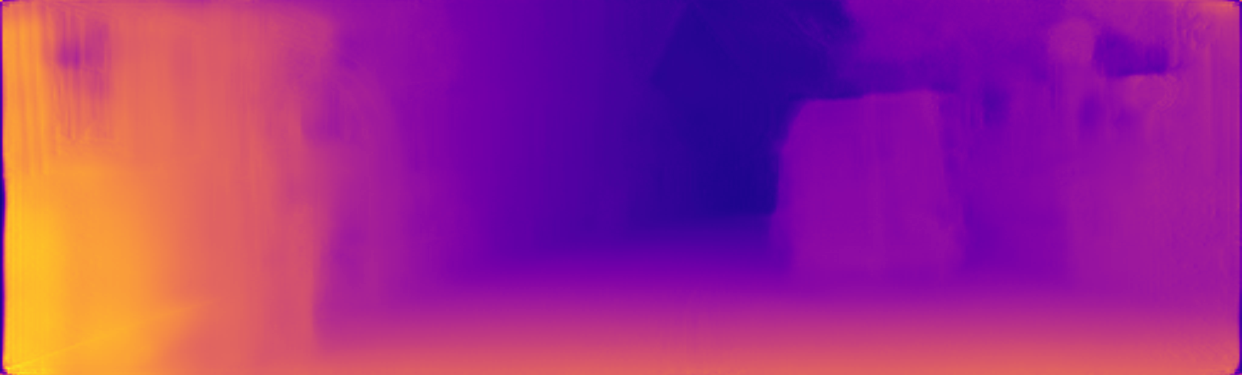}
      &\includegraphics[width=0.235\textwidth, valign=c]{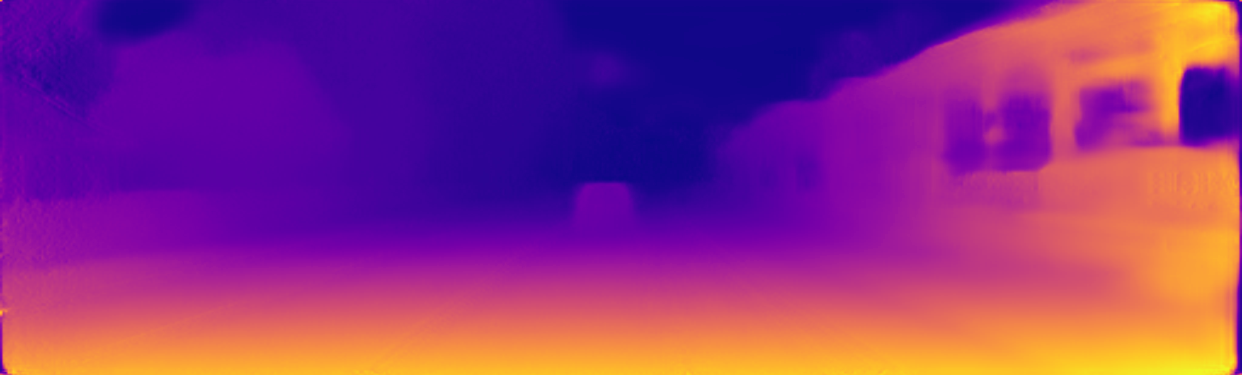}
      &\includegraphics[width=0.235\textwidth, valign=c]{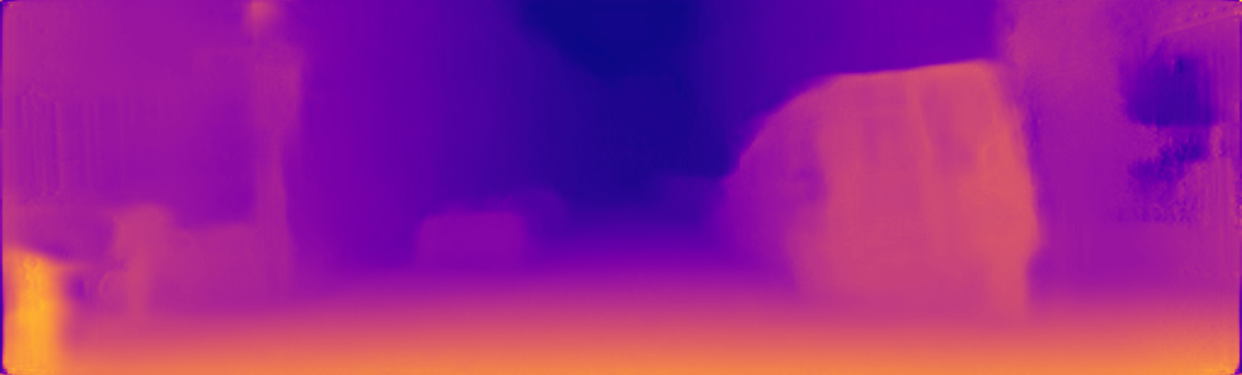}
      &\includegraphics[width=0.235\textwidth, valign=c]{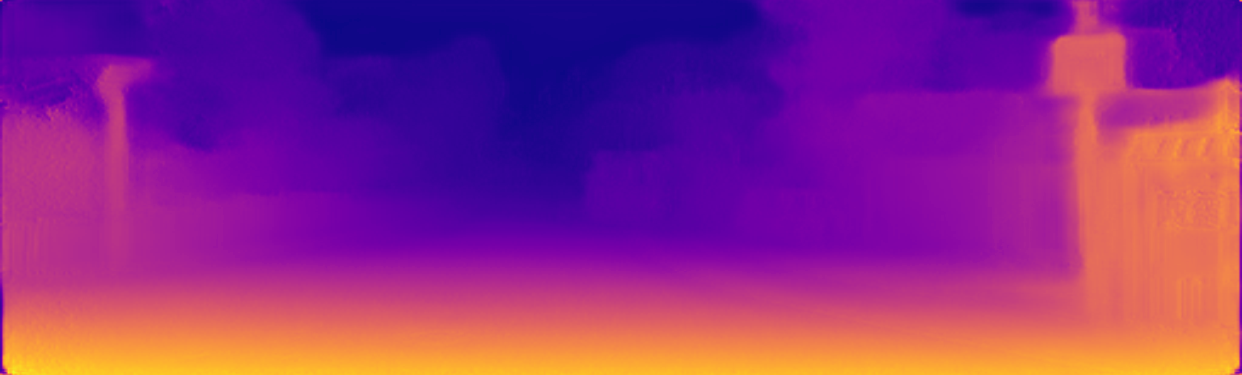}\\
      \rotatebox[origin=c]{90}{Monodepth2(S)\cite{godard2019digging}}
      &\includegraphics[width=0.235\textwidth, valign=c]{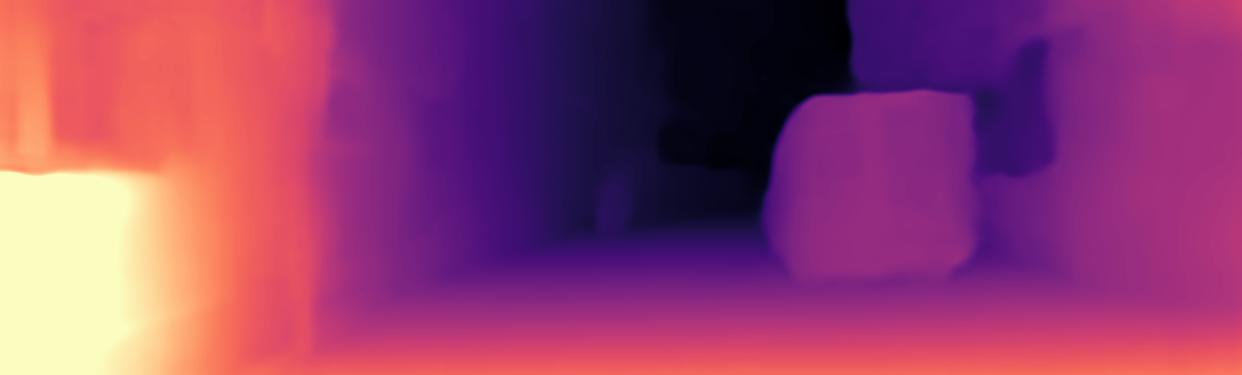}
      &\includegraphics[width=0.235\textwidth, valign=c]{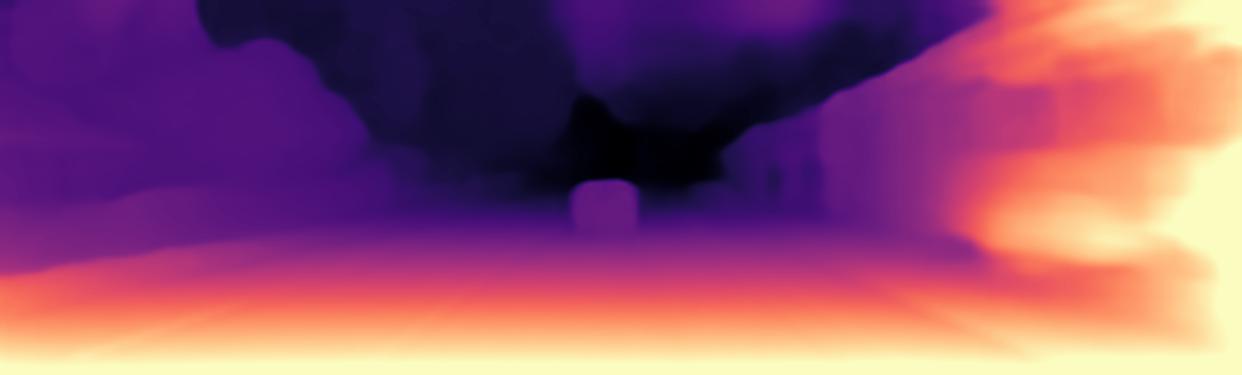}
      &\includegraphics[width=0.235\textwidth, valign=c]{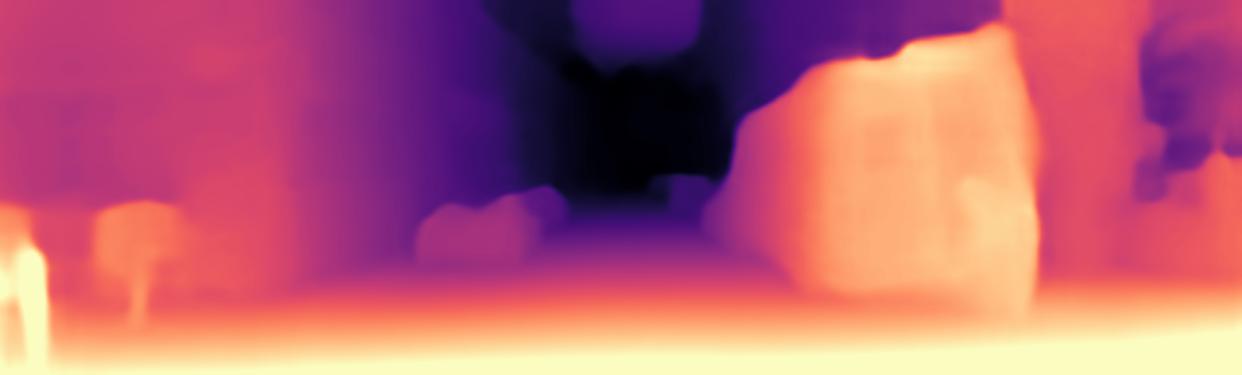}
      &\includegraphics[width=0.235\textwidth, valign=c]{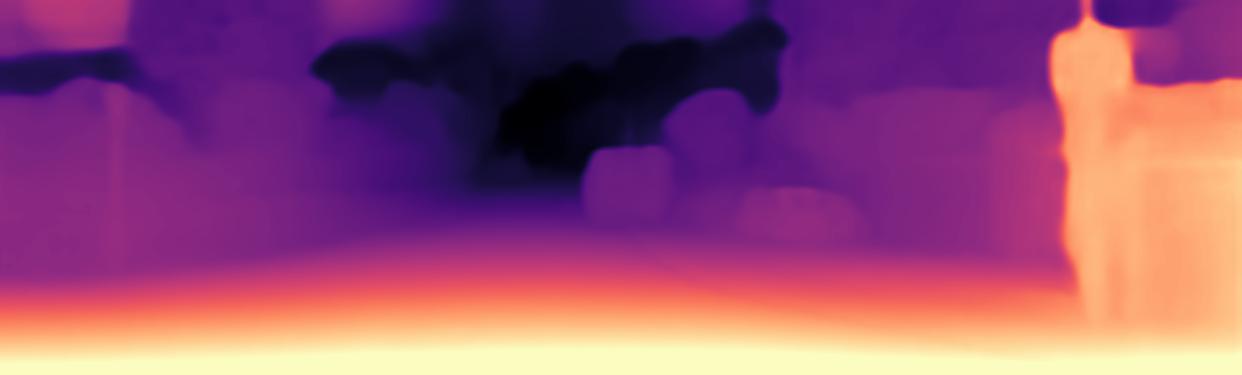}\\
      \rotatebox[origin=c]{90}{Pilzer\cite{pilzer2019progressive}}
      &\includegraphics[width=0.235\textwidth, valign=c]{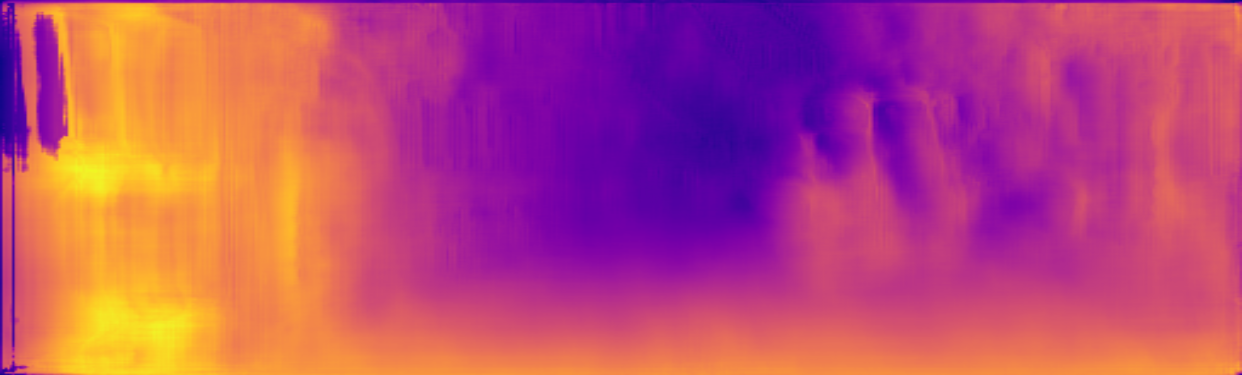}
      &\includegraphics[width=0.235\textwidth, valign=c]{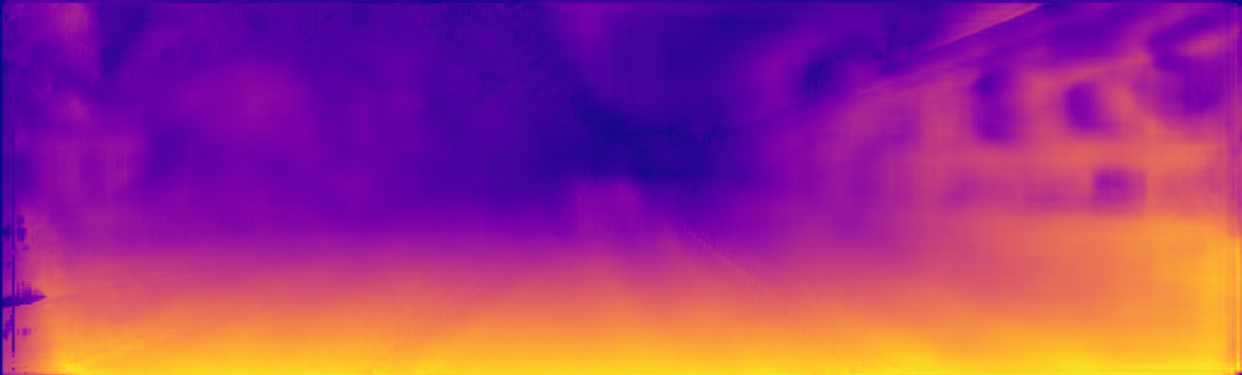}
      &\includegraphics[width=0.235\textwidth, valign=c]{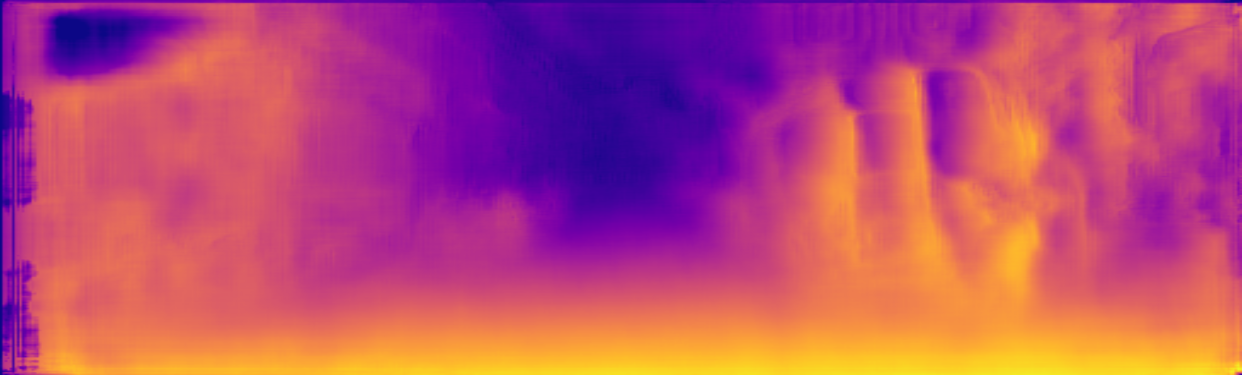}
      &\includegraphics[width=0.235\textwidth, valign=c]{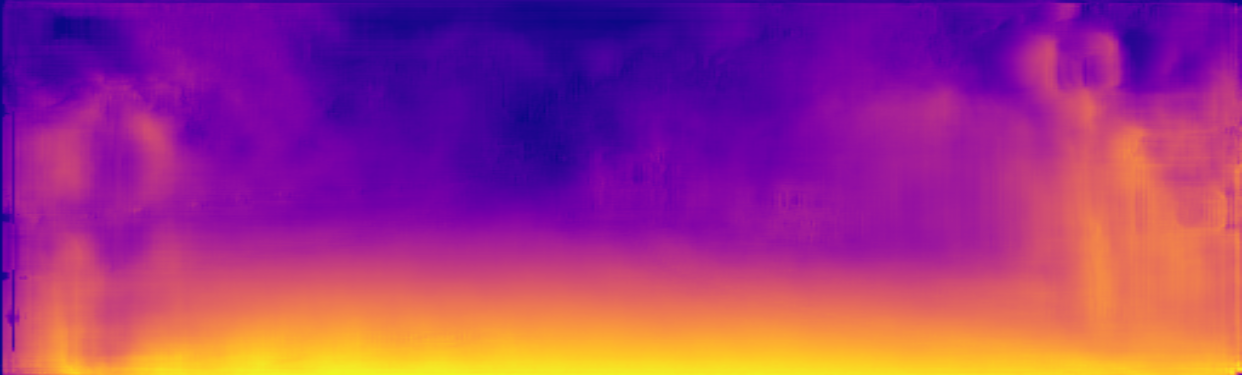}
      \vspace{1mm}\\
      \rotatebox[origin=c]{90}{\textbf{H-NET(ours)}}
      &\includegraphics[width=0.235\textwidth, valign=c]{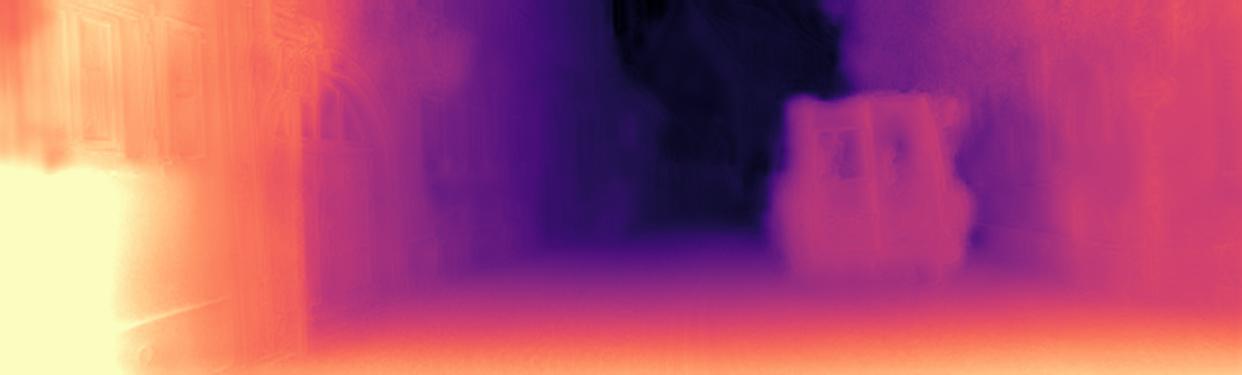}
      &\includegraphics[width=0.235\textwidth, valign=c]{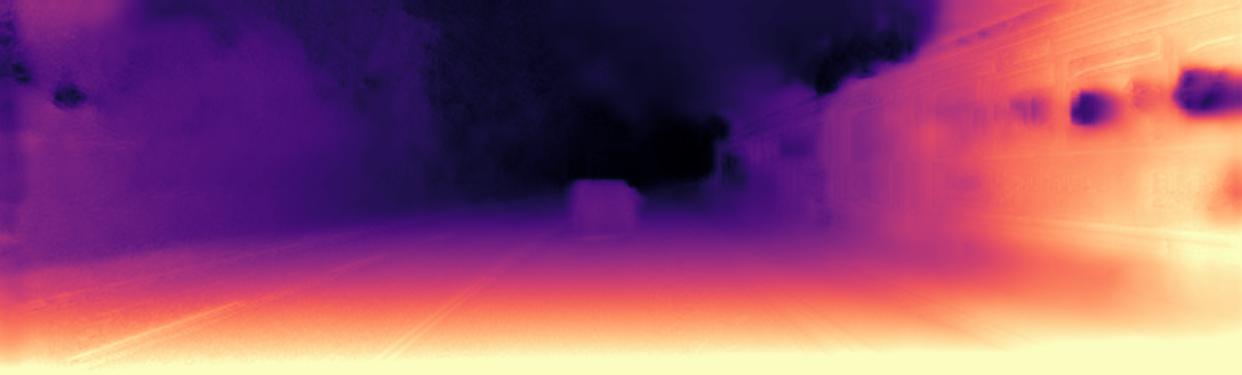}
      &\includegraphics[width=0.235\textwidth, valign=c]{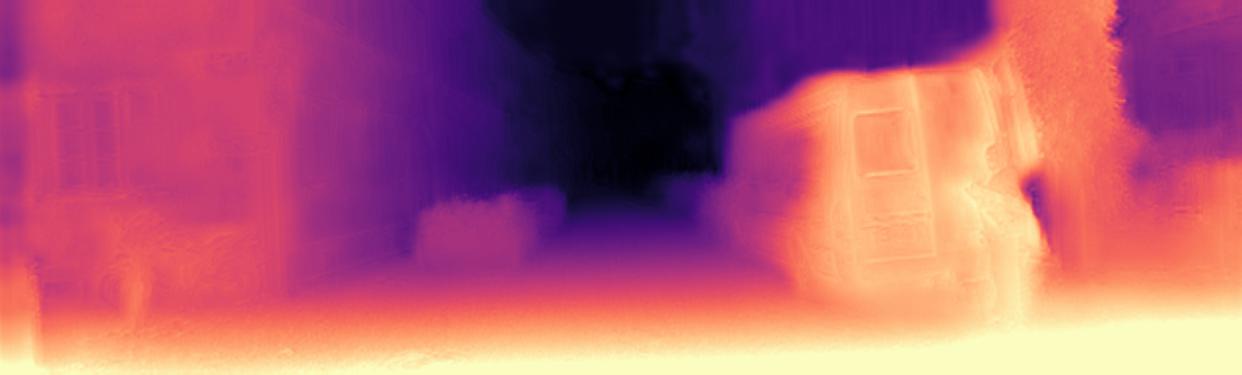}
      &\includegraphics[width=0.235\textwidth, valign=c]{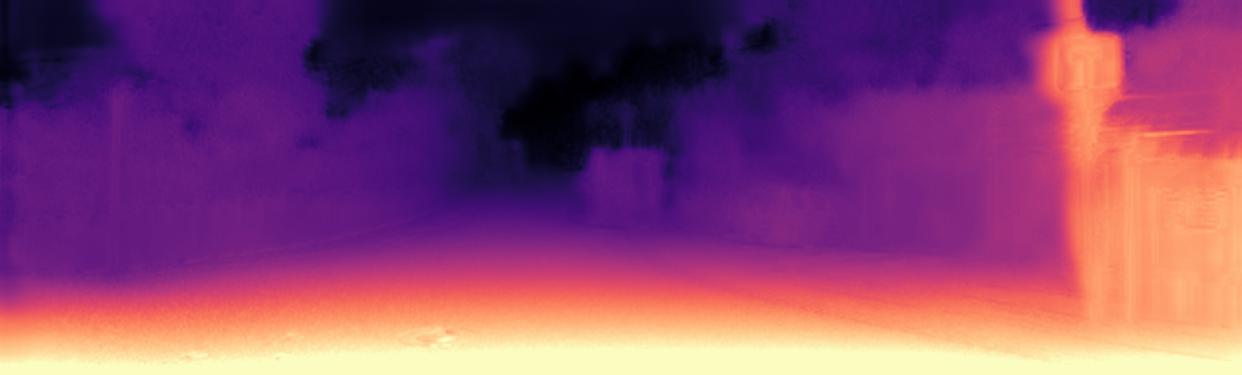}\\
  \end{tabular}
  \vspace{0.6mm}
  \caption{Qualitative results on the KITTI Eigen split. The depth prediction are all for the left input image. Our H-Net in the last row generates the depth maps with more details and performs better on distinguishing different parts in one object, \textit{i.e.} buildings, kerbs bushes and trees, which reflects the superior quantitative results in Table \ref{quantitive results}.}
  \label{Fig:final_result}
  \end{figure*}


\begin{table*}[]

\resizebox{\textwidth}{!}{
\begin{tabular}{|c|c|c|c||c|c|c|c|c|c|c|}
\hline

\hline
Setting & SE-SD &MEA &OT
& \cellcolor[RGB]{255,170,170}Abs Rel
& \cellcolor[RGB]{255,170,170}Sq Rel 
& \cellcolor[RGB]{255,170,170}RMSE  
& \cellcolor[RGB]{255,170,170}RMSE log 
& \cellcolor[RGB]{153,204,255}${\delta < 1.25 }$ &\cellcolor[RGB]{153,204,255} $\delta < 1.25^2$  &\cellcolor[RGB]{153,204,255}${\delta < 1.25^3}$ \\ 
\hline

\hline
Backbone \cite{godard2019digging} &\ding{55}  &\ding{55} &\ding{55} &0.109 &0.873 &4.960 &0.209 &0.864 &0.948 &0.975   \\

SE-SD &\checkmark &\ding{55}  &\ding{55}  &0.096  &0.700   &4.403  &0.189   &0.894  &0.960   &0.979  \\
SE-SD w/ EG-MNL  &\checkmark &\ding{55}  &\ding{55} & {0.086}   & 0.701  & 4.289 & 0.178   & {0.912}  & {0.964}  &0.980 \\ 
SE-SD w/ EG-MEA &\checkmark  &\checkmark &\ding{55}  &0.080 &0.665 &4.086 &0.173  & 0.917  &0.964  &0.981 \\
SE-SD w/ OT-MNL  &\checkmark  &\ding{55} &\checkmark   &0.082  &0.725  &4.279  &0.180  &0.917  &0.962  &0.979\\
\rowcolor[RGB]{230,230,230} \textbf{H-Net (Ours)} &\checkmark  &\checkmark &\checkmark  &\textbf{0.076}  &\textbf{0.607}  &\textbf{4.025}  & \textbf{0.166}  &\textbf{0.918}  &\textbf{0.966}  &\textbf{0.982}   \\
\hline

\hline
\end{tabular}}
\vspace{0.5mm}
\caption{Ablation Study. Results for different variants of our model (H-Net) on KITTI2015 \cite{geiger2012we} using full Eigen dataset with comparison to our backbone Monodepth2 \cite{godard2019digging}. We evaluate the impact of the Siamese encoder- Siamese decoder (SE-SD), mutual epipolar attention (MEA) and optimal transport (OT). Metrics labeled by red mean \textit{lower is better} while labeled by blue mean \textit{higher is better}}
\label{Abalation study}
\end{table*}


\begin{table}
\begin{center}
\begin{tabular}{|c|c|}
\hline

\hline
Setting & Num of Parameters\\
\hline
SE-SD (baseline)  &30.7M   \\
\hline
EG-MNL   &+0.3M(1\%)  \\
EG-MEA   &+0.6M(2\%)  \\
OT-MNL    &+0.3M(1\%) \\
\rowcolor[RGB]{230,230,230} \textbf{OT-MEA (Ours)}   &+0.6M(2\%) \\
\hline

\end{tabular}
\end{center}
\caption{Number of Parameters (M:million) for our models with different settings of the mutual attention module.}
\label{parameters}
\end{table}

\begin{figure*}[ht!]
  \centering
  \centering
  \label{Fig:final_result}
  \tiny
  \setlength{\tabcolsep}{0.18em}
  \begin{tabular}{ccccc}
      \rotatebox[origin=c]{90}{Left input} 
      &\includegraphics[width=0.235\textwidth, valign=c]{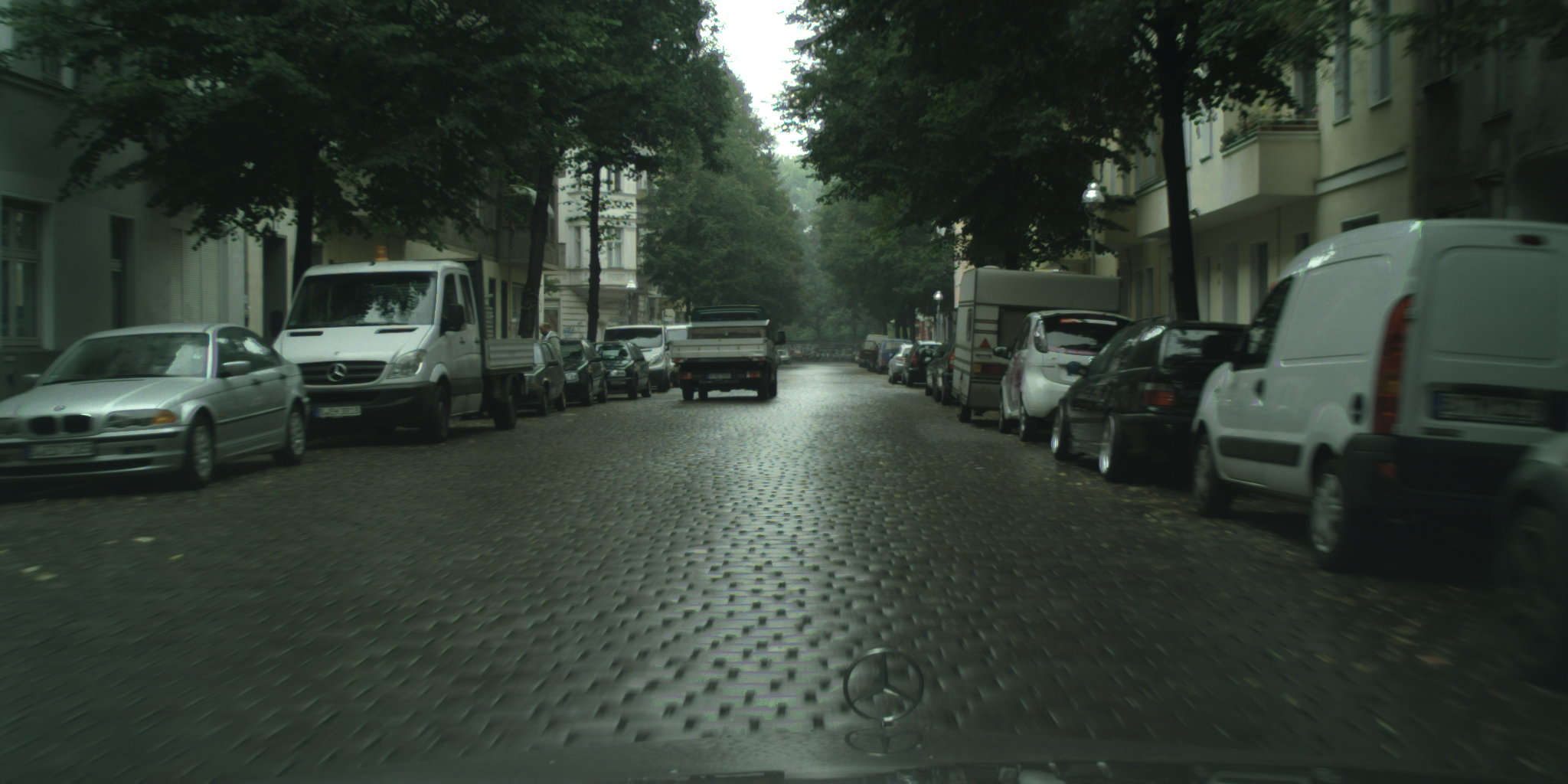}
      &\includegraphics[width=0.235\textwidth, valign=c]{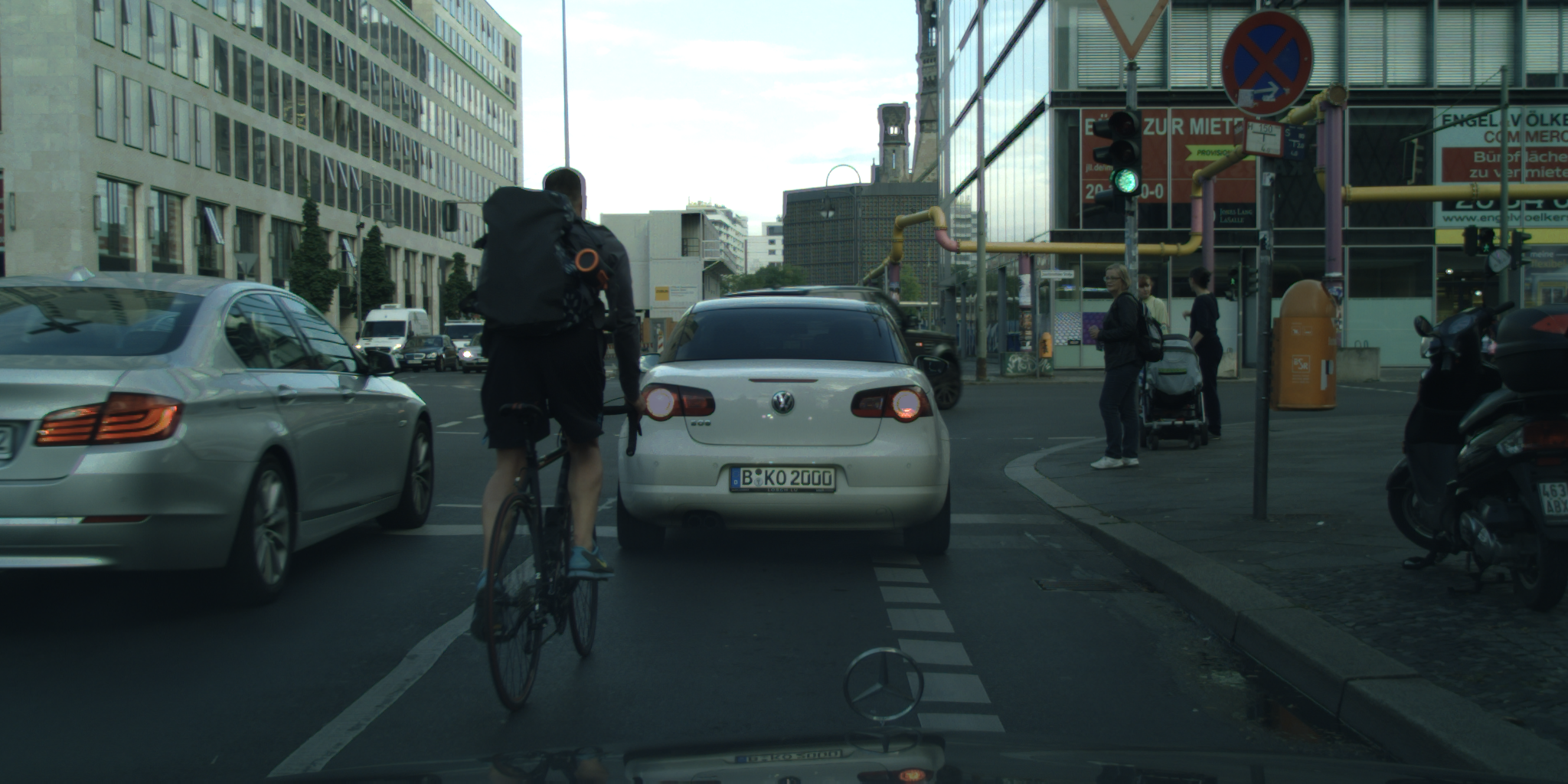}
      &\includegraphics[width=0.235\textwidth, valign=c]{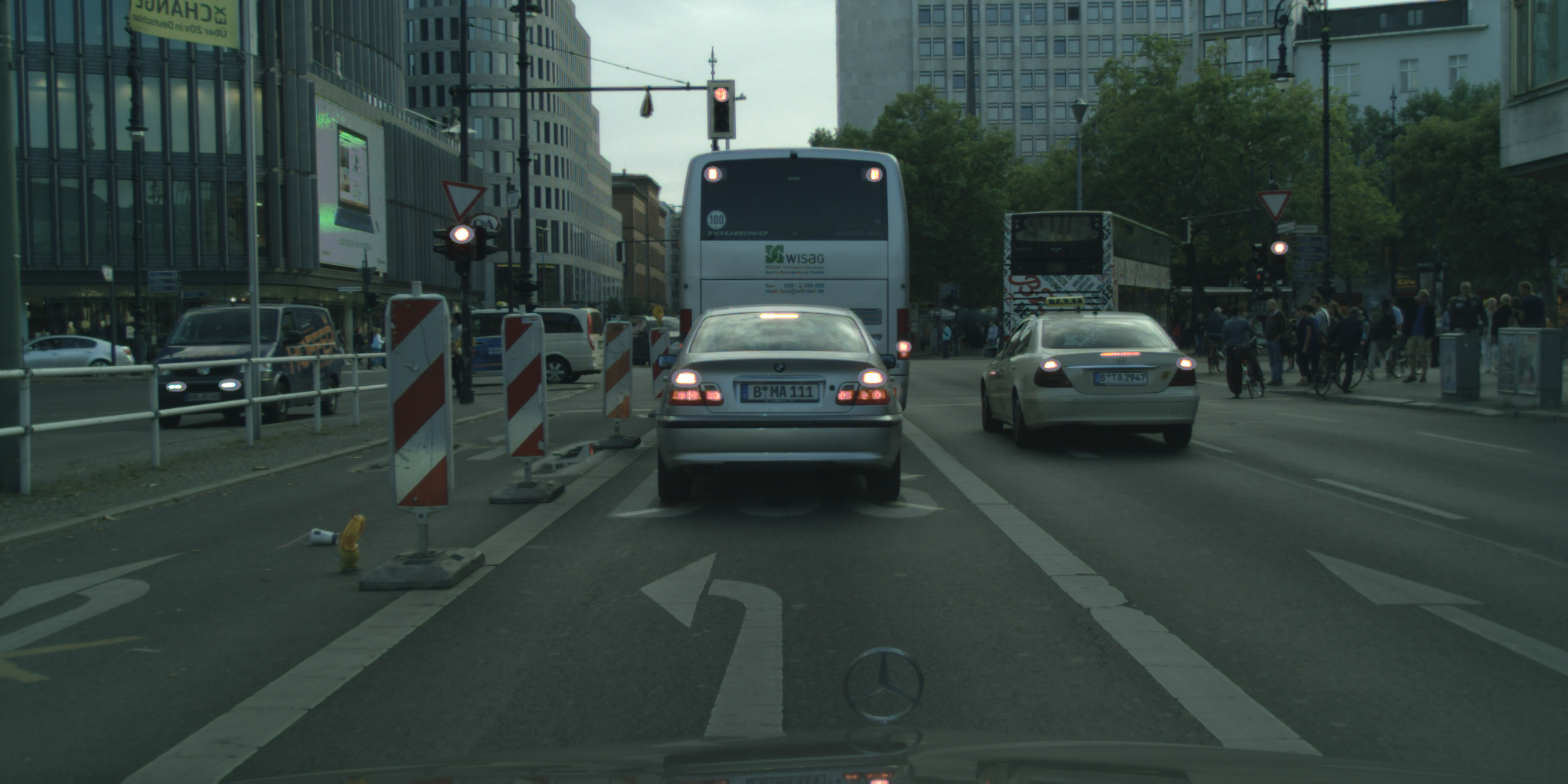}
      &\includegraphics[width=0.235\textwidth, valign=c]{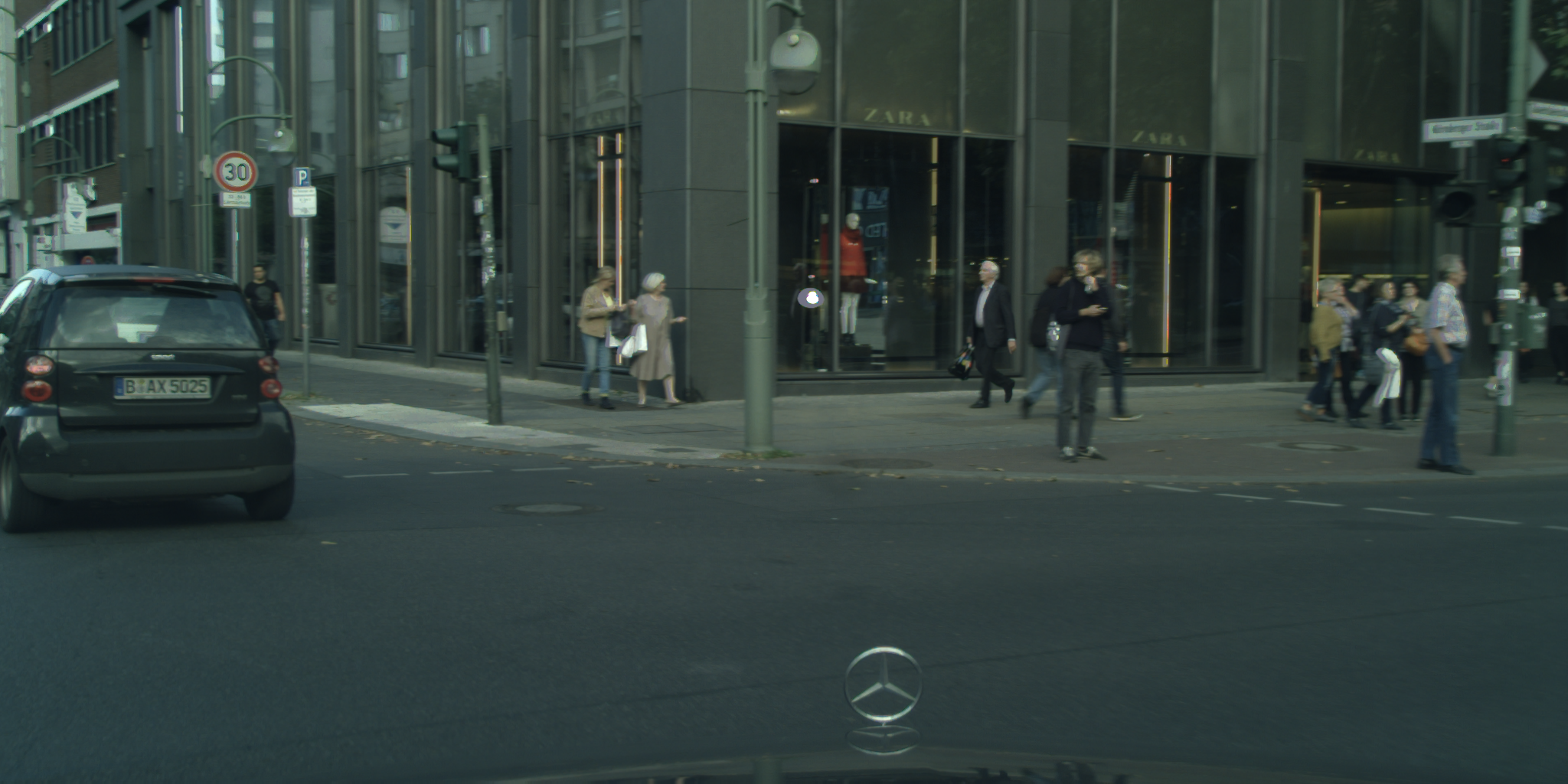}
      \vspace{0.2mm}\\
      \rotatebox[origin=c]{90}{Right input}
      &\includegraphics[width=0.235\textwidth, valign=c]{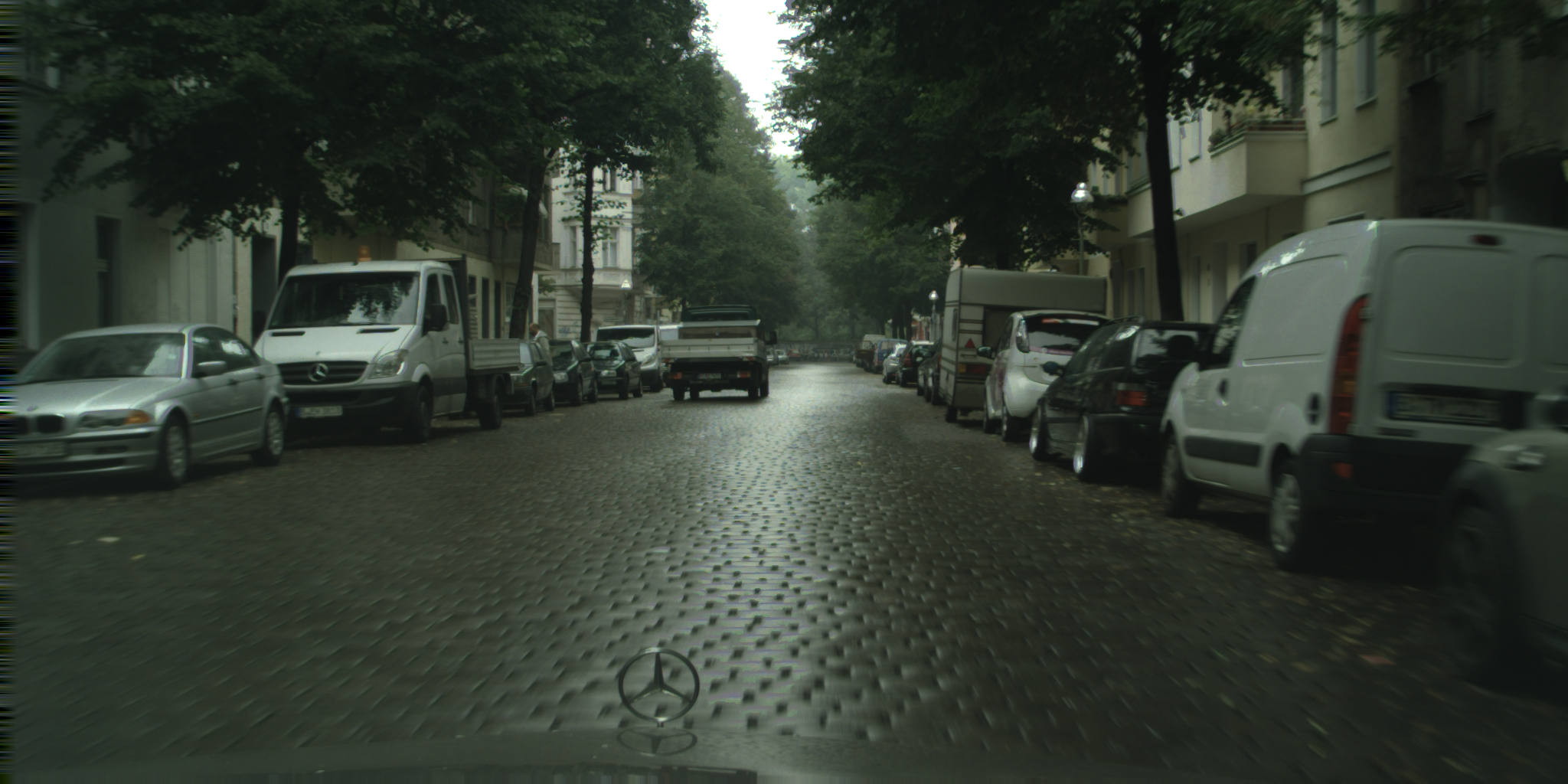}
      &\includegraphics[width=0.235\textwidth, valign=c]{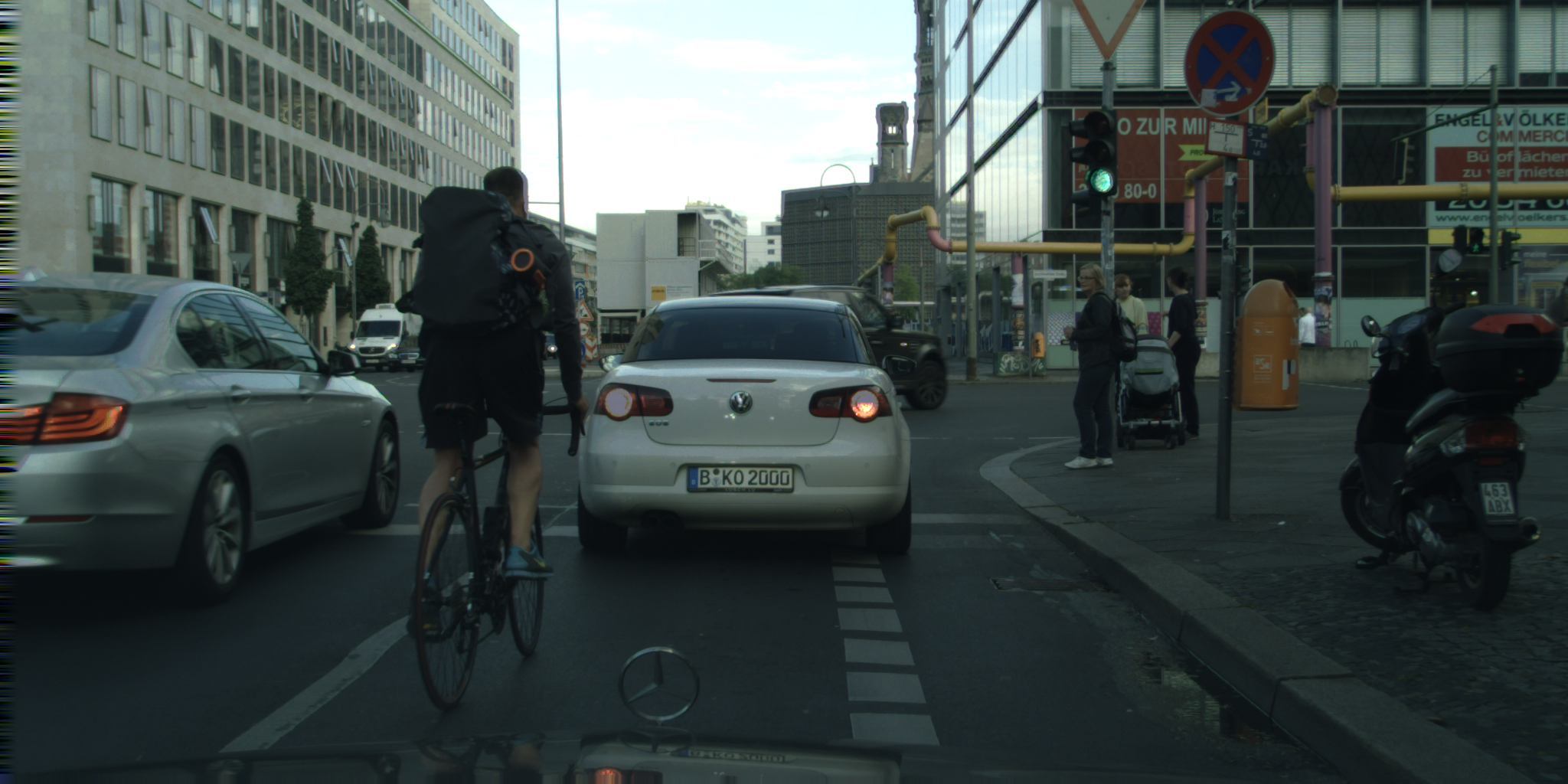}
      &\includegraphics[width=0.235\textwidth, valign=c]{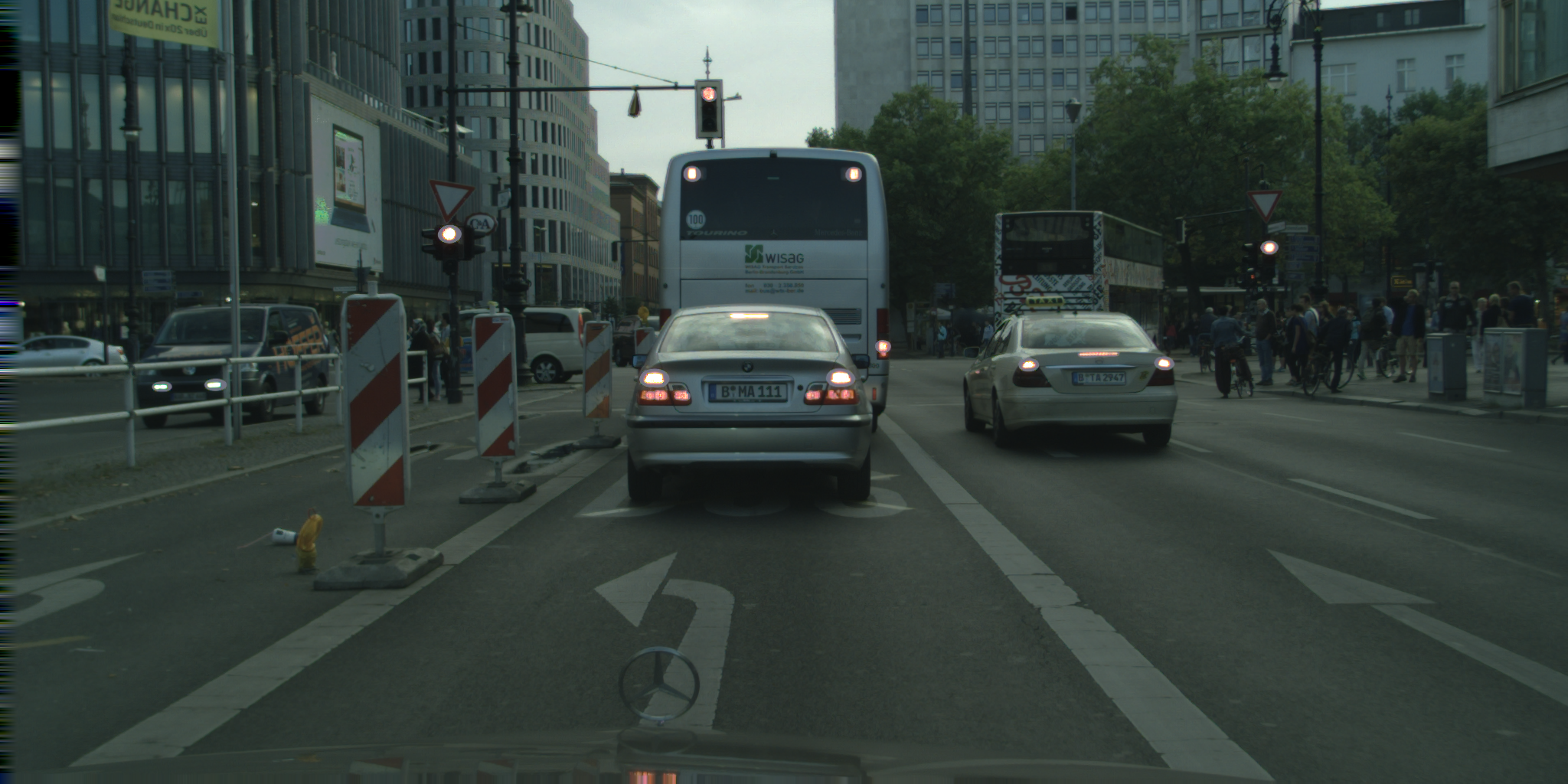}
      &\includegraphics[width=0.235\textwidth, valign=c]{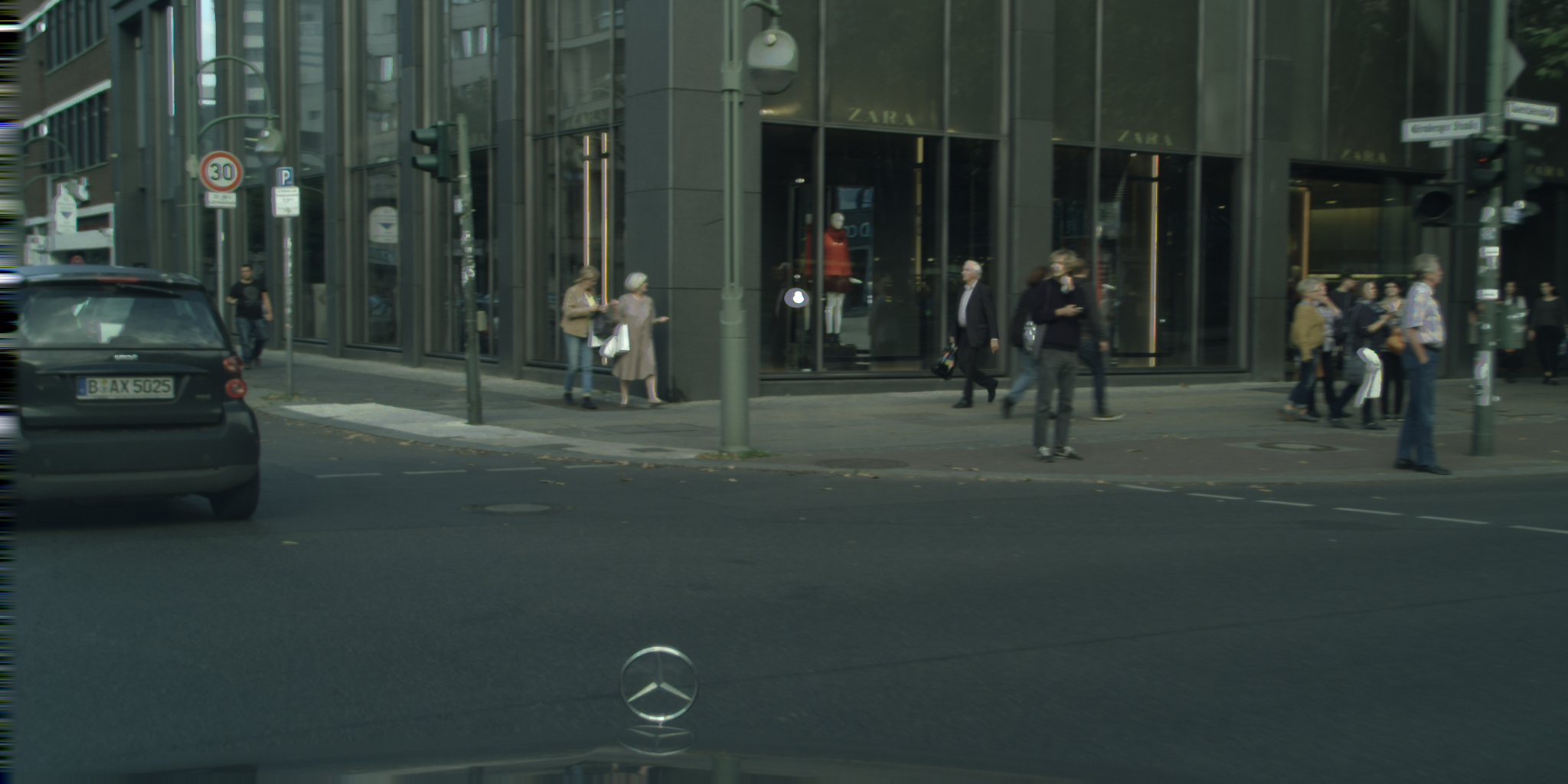}
      \vspace{0.2mm}\\
      \rotatebox[origin=c]{90}{\textbf{H-Net}}
      &\includegraphics[width=0.235\textwidth, valign=c]{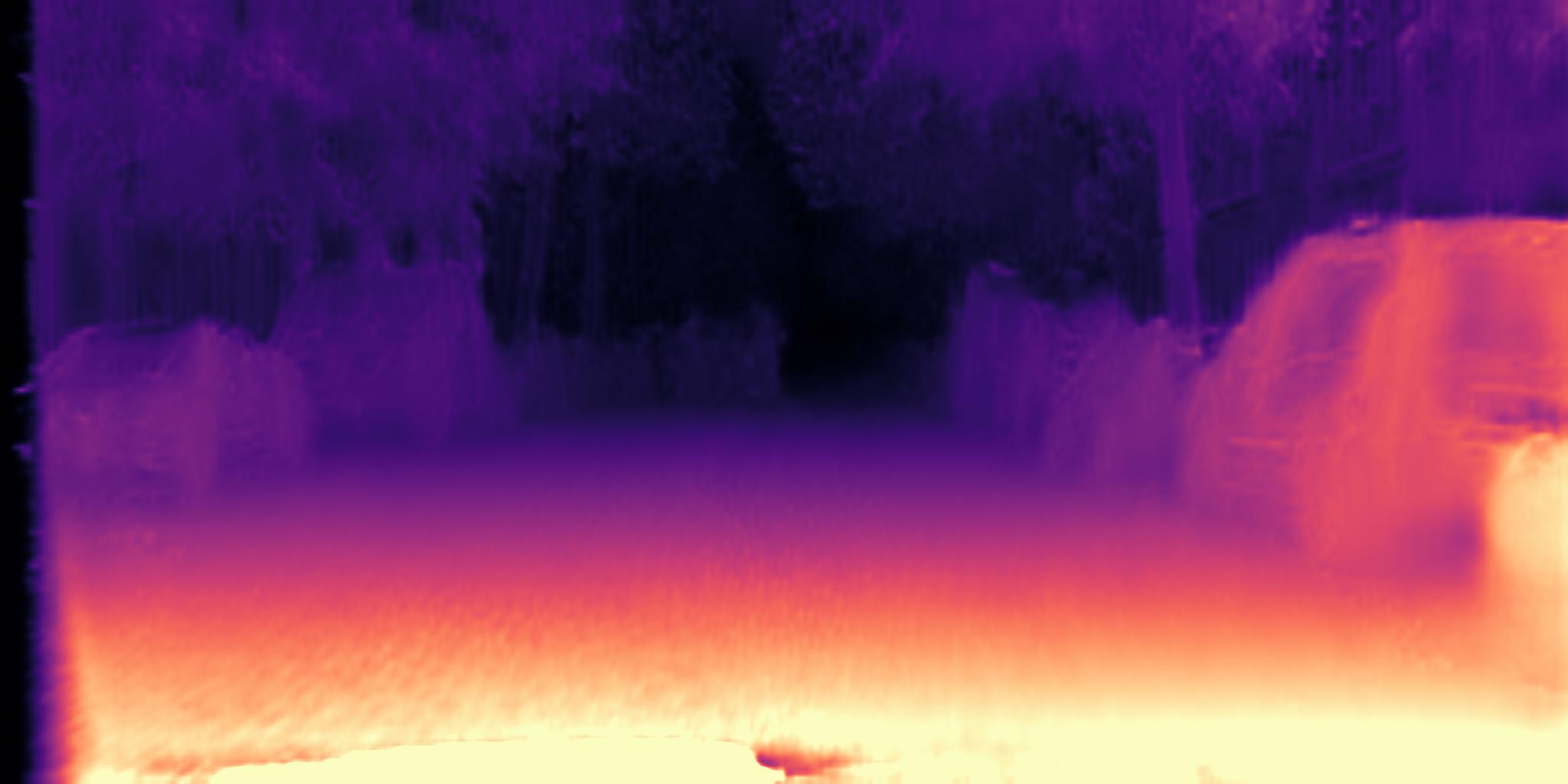}
      &\includegraphics[width=0.235\textwidth, valign=c]{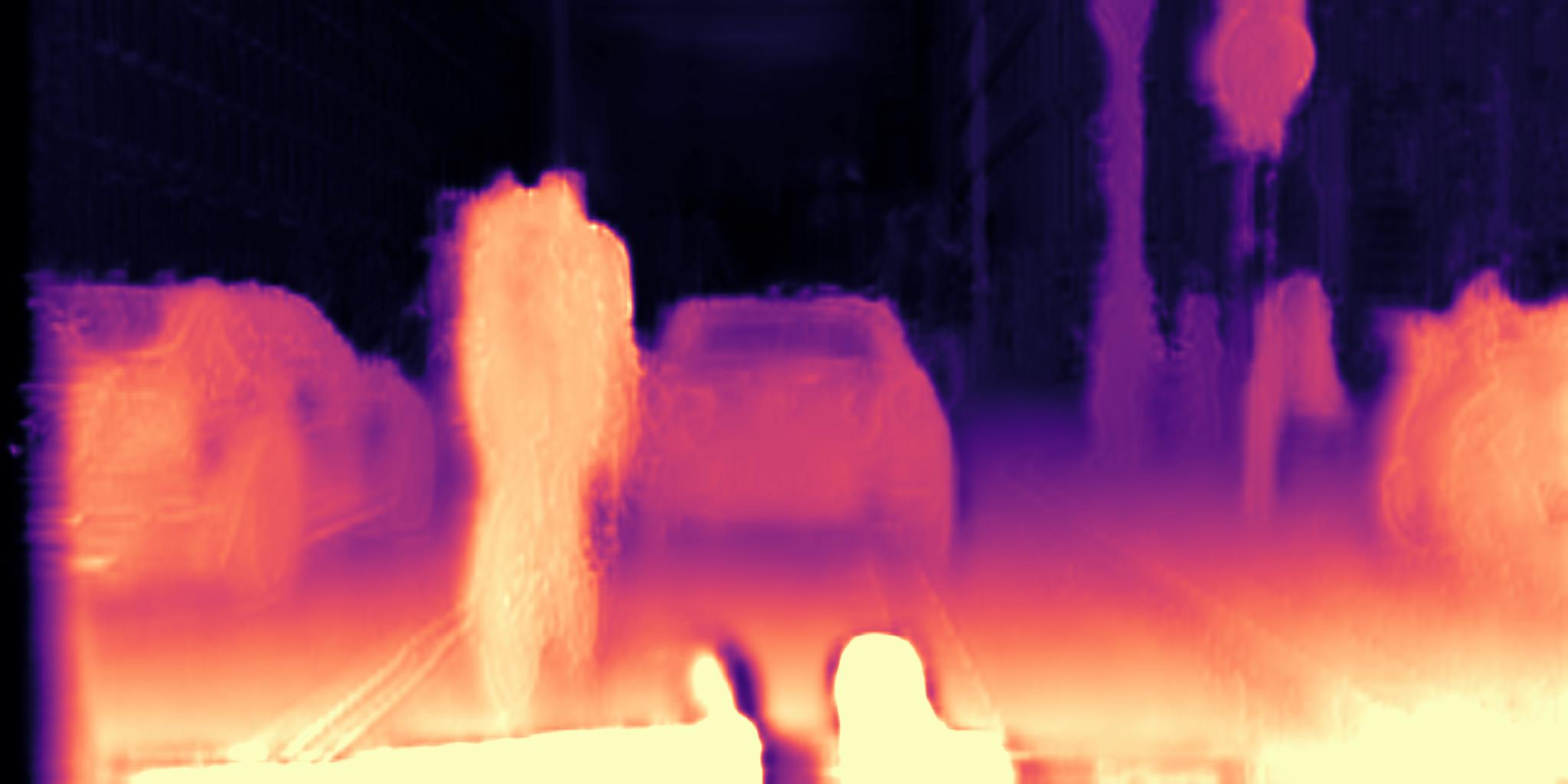}
      &\includegraphics[width=0.235\textwidth, valign=c]{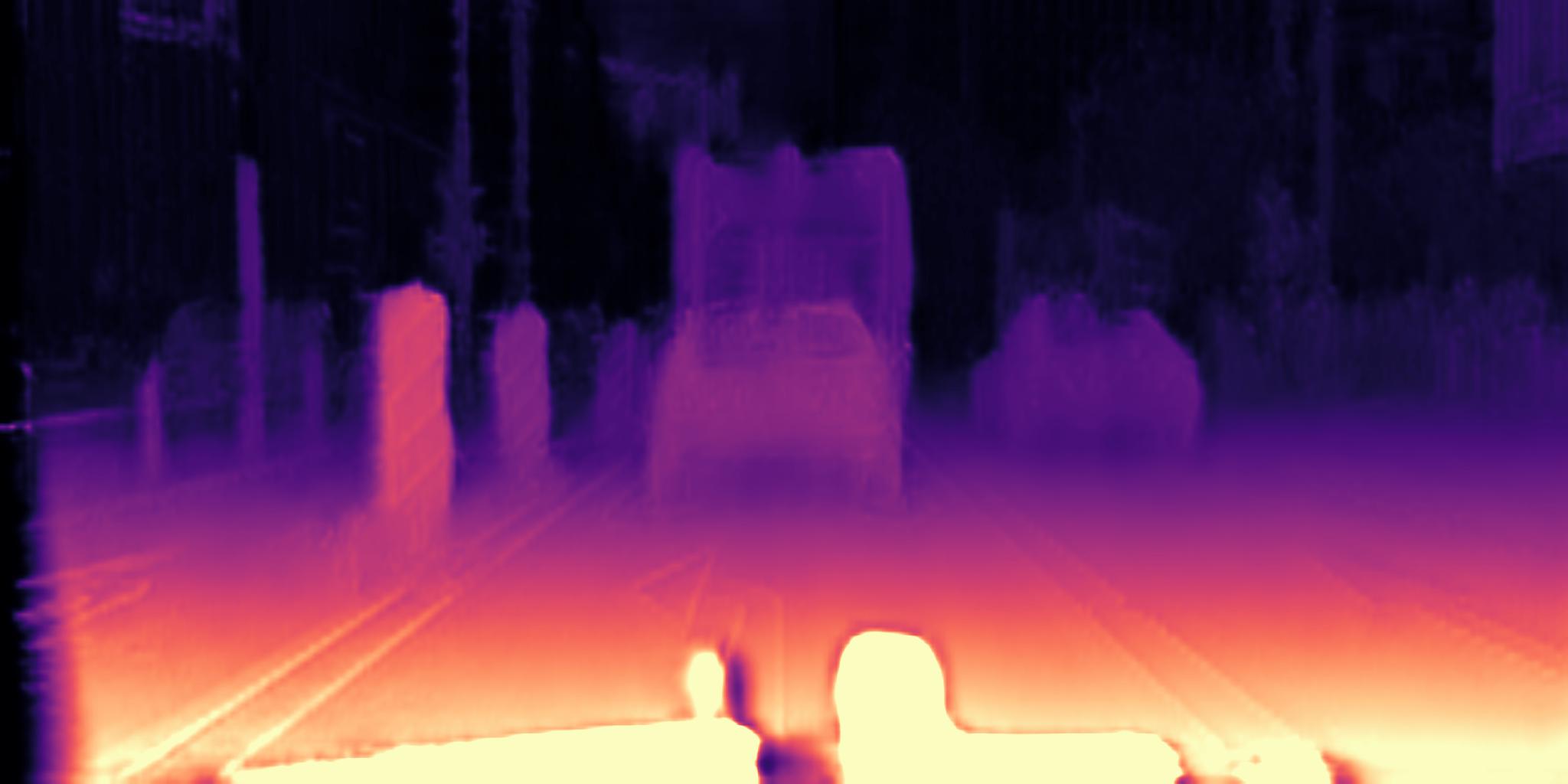}
      &\includegraphics[width=0.235\textwidth, valign=c]{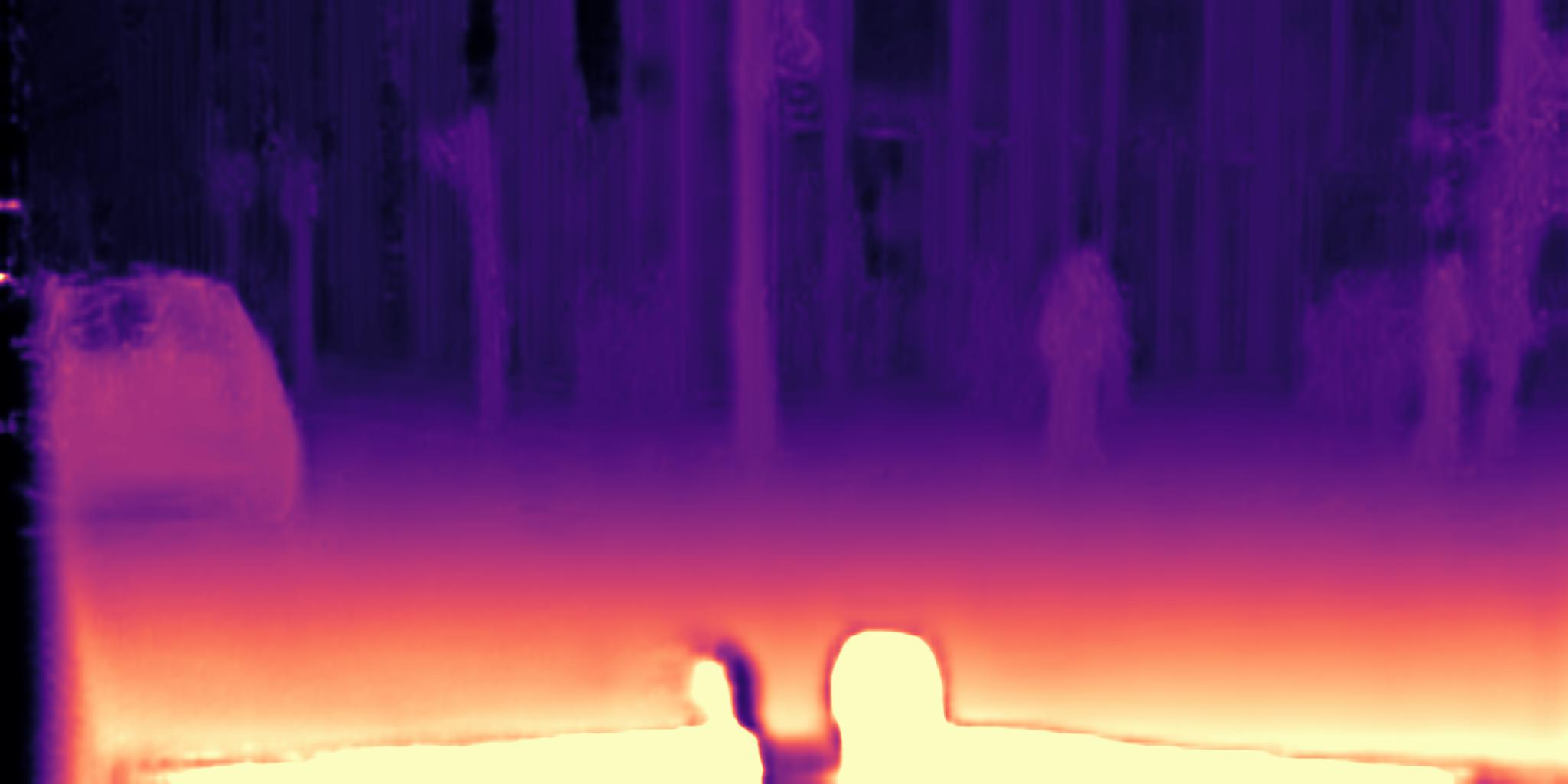}
      \vspace{0.5mm}\\
      \rotatebox[origin=c]{90}{Ground Truth}
      &\includegraphics[width=0.235\textwidth, valign=c]{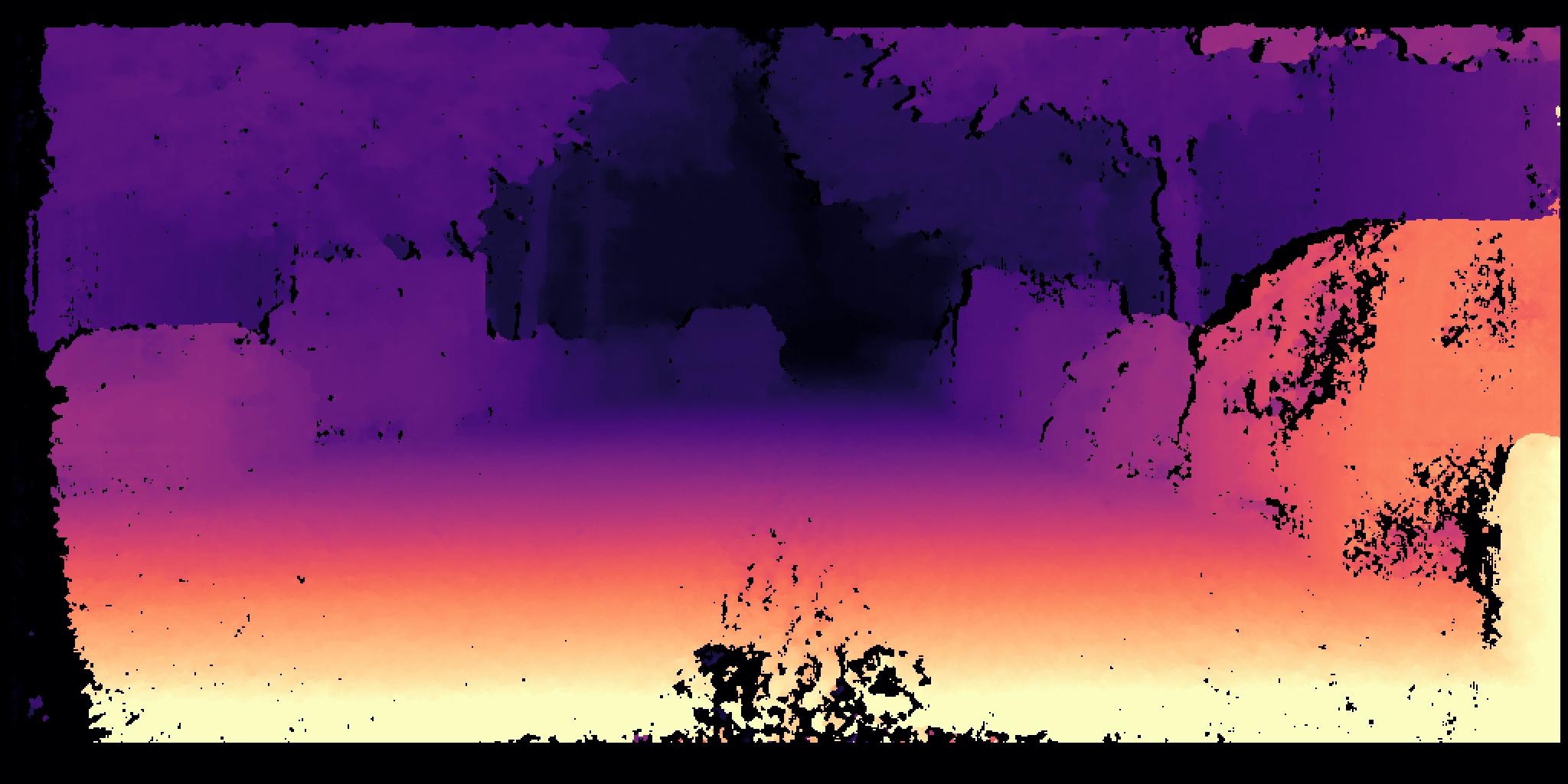}
      &\includegraphics[width=0.235\textwidth, valign=c]{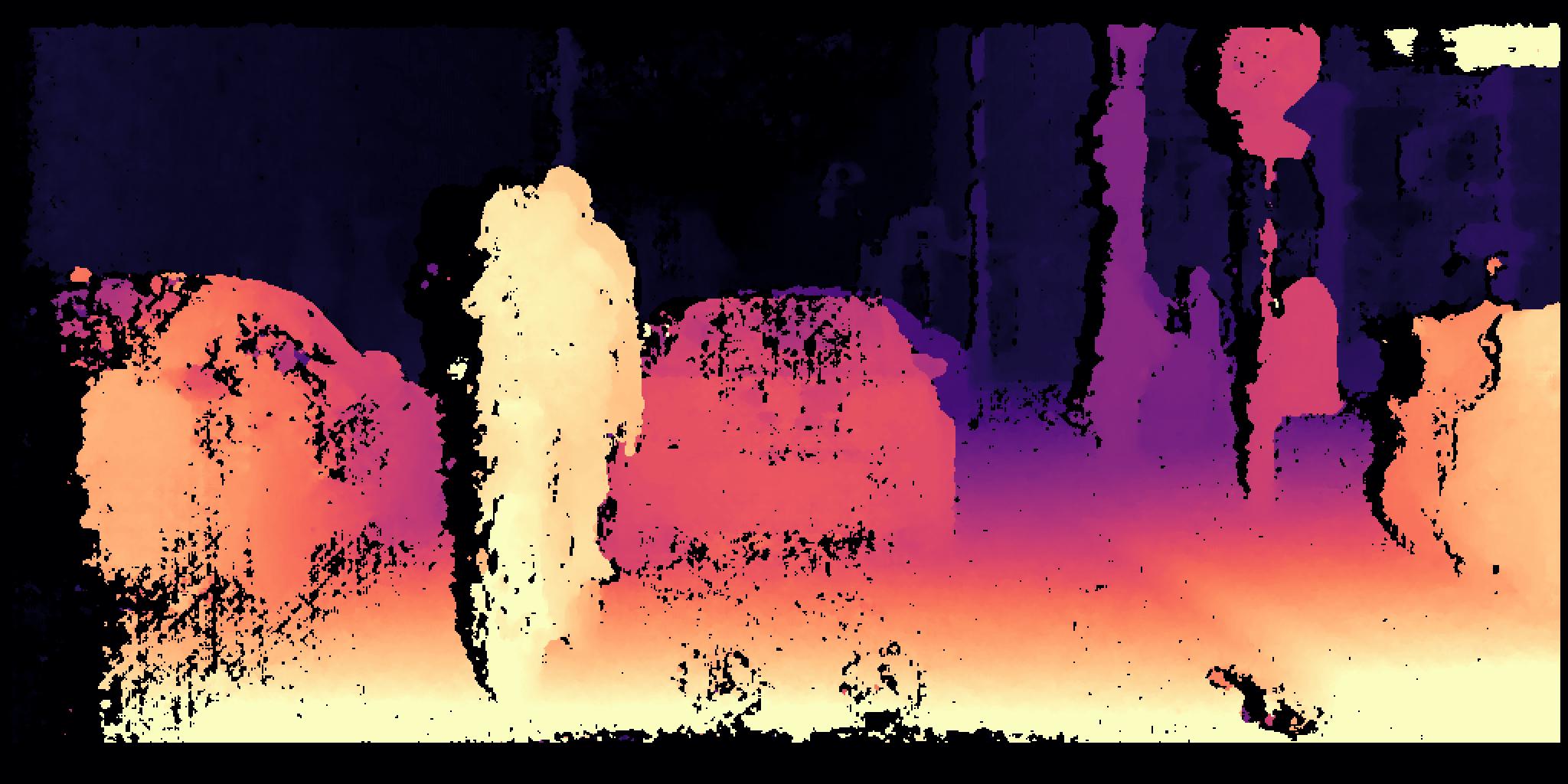}
      &\includegraphics[width=0.235\textwidth, valign=c]{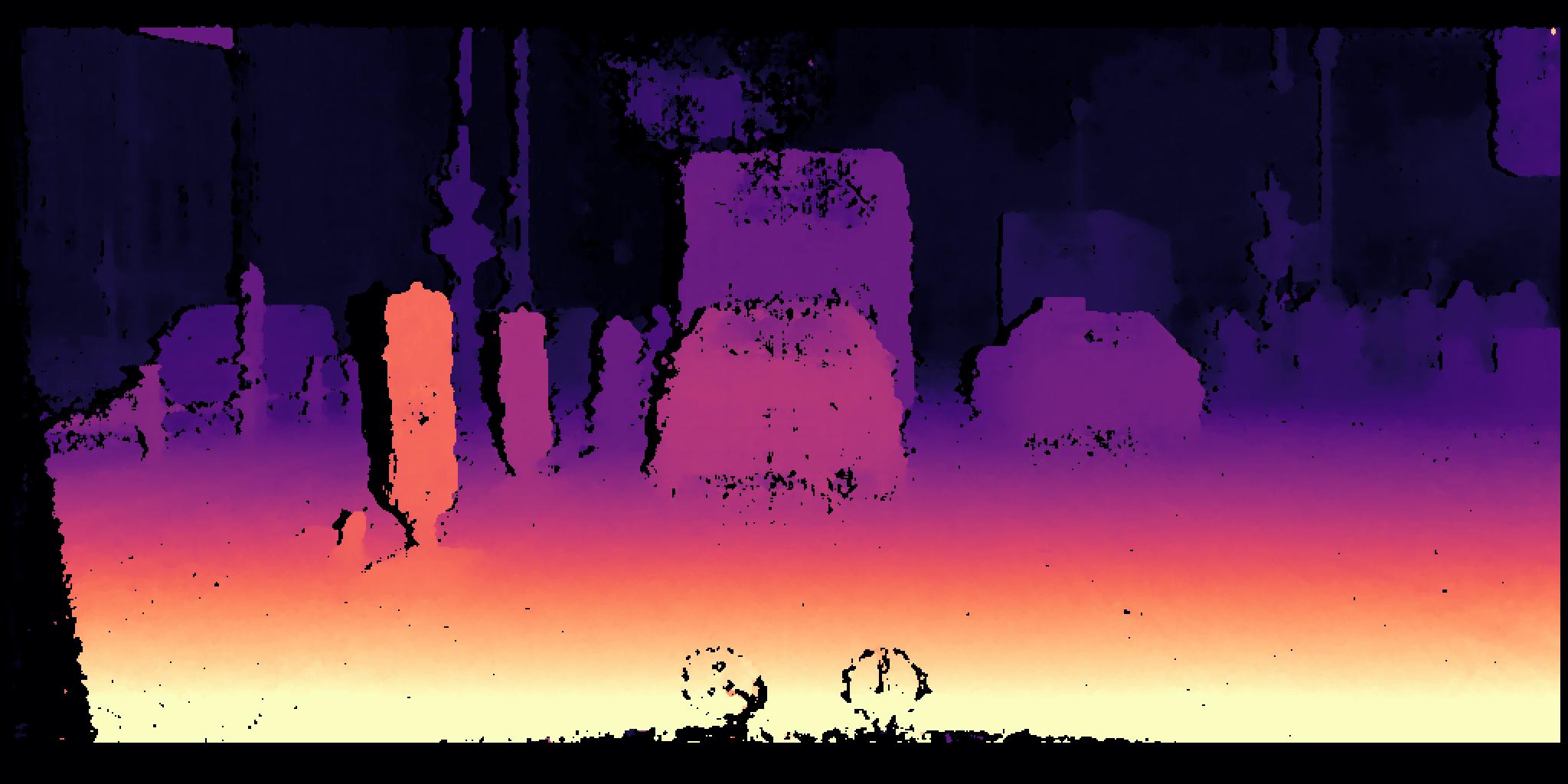}
      &\includegraphics[width=0.235\textwidth, valign=c]{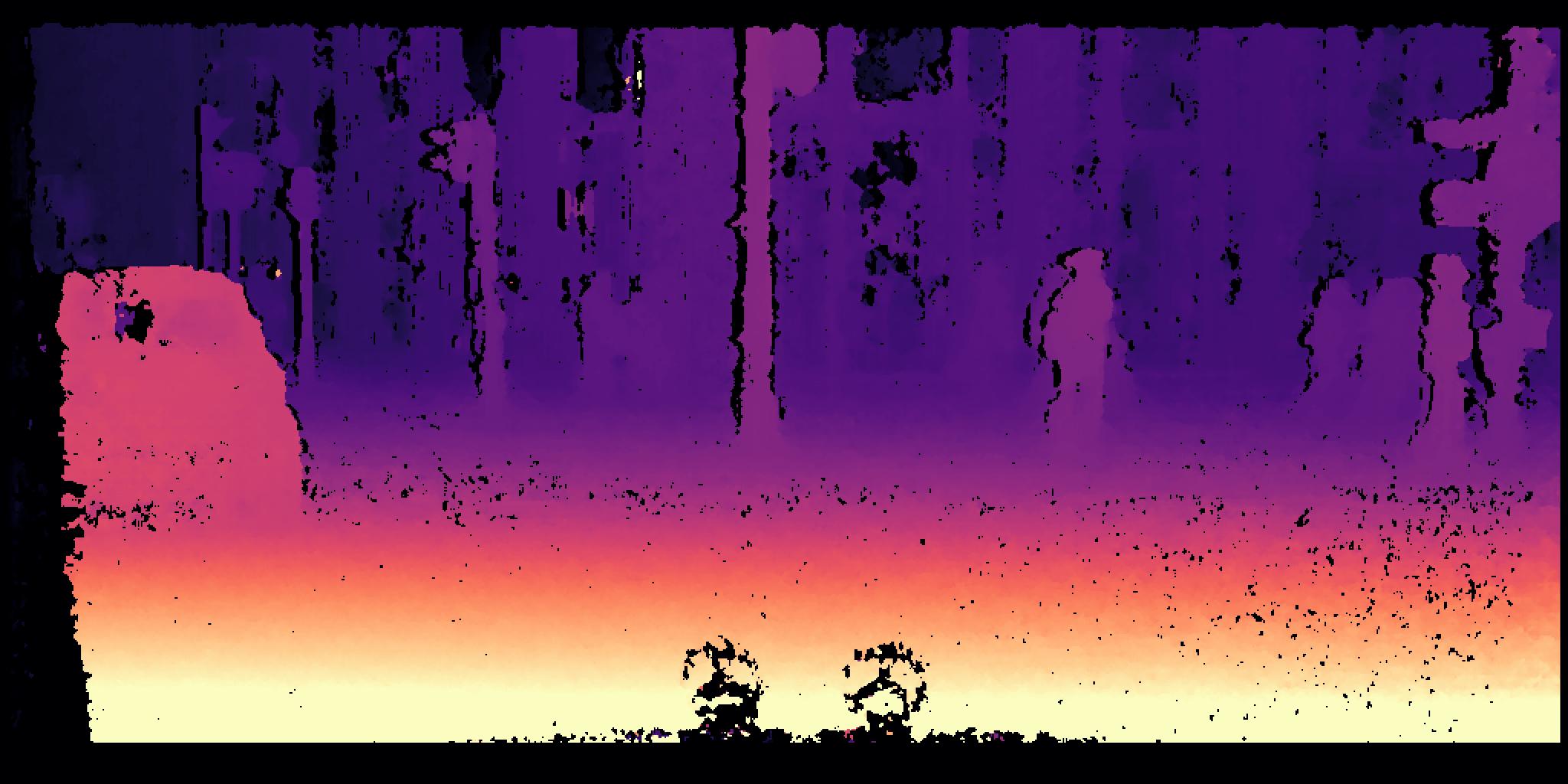}\\
     
  \end{tabular}
  \vspace{0.5mm}
  \caption{Qualitative results on the Cityscapes dataset. Our H-Net generates very close predictions compared with the ground truth.}
  \label{Fig:cityscape}
  \end{figure*}

We trained and evaluated the proposed H-Net on the KITTI2015 \cite{geiger2012we} with the full Eigen and Eigen split dataset \cite{eigen2014depth}. For the full Eigen setting, there were 22600 pairs for training and 888 for validation while for Eigen split, there were 19905 training pairs and 2212 validation pairs. The same intrinsics were used for all images. The principal point of the camera was set to the image center and the focal length was the average of all the focal lengths in KITTI. All of the images were rectified and the transformation between the two stereo images was set to be a pure horizontal translation of fixed length. During the evaluation, only depths up to a fixed range of 80m were evaluated per standard practice \cite{eigen2014depth, garg2016unsupervised, Godard_2017_CVPR, godard2019digging}. As our backbone model, we used Monodepth2 \cite{godard2019digging} and kept the original ResNet18 \cite{he2016deep} as the encoder. Furthermore, we also trained and tested our H-Net on the Cityscapes dataset \cite{cordts2016cityscapes} to verify its generalisability.  

We compared our results with state-of-the-art supervised and self-supervised approaches and both qualitative and quantitative results were generated for comparison. To better understand how each component influenced the overall performance, we conducted an ablation study by turning various components of the model off,  in turn. 

\subsection{Implementation Details}
Our H-Net model was trained using the PyTorch library \cite{paszke2017automatic}, with an input/output resolution of $640\times 192$ and a batch size of 8. The \textit{$L_1$}-norm loss term \(\gamma\) was set to 0.85 and the smoothness term $\lambda$ was 0.001. The number of scales $m$ was set to 4, which meant that totally we had 4 multi scales and there were 4 output scales as well with resolutions \(\frac{1}{2^0}\), \(\frac{1}{2^1}\), \(\frac{1}{2^2}\) and \(\frac{1}{2^3}\) of the input resolution. The model was trained for 20 epochs using the Adam optimizer \cite{kingma2014adam} requiring approximately 14 hours on a single NVIDIA 2080Ti GPU. We set the learning rate to $10 ^{-4}$ for the first 15 epochs and dropped it to $10^{-5}$ for the remainder. As with previous papers \cite{godard2019digging}, we also used a Resnet encoder with pre-trained weights on ImageNet \cite{russakovsky2015imagenet}, which proved able to improve the overall accuracy of the depth estimation and to reduce the training time \cite{godard2019digging,johnston2020self}.

\section{Results and Discussion}






\subsection{KITTI Results}
The qualitative results and quantitive results on the KITTI Eigen split are shown in Table \ref{quantitive results} and Figure \ref{Fig:final_result}. In Table~\ref{quantitive results}, it can be seen that the proposed H-Net outperforms all existing state-of-the-art self-supervised methods by a significant margin. Compared with other approaches that applied direct supervision signals (supervised methods), the model was still competitive. As our H-Net takes stereo image pairs as the input, in contrast with \cite{godard2019digging} and \cite{johnston2020self}, we did not need to remove static frames. However, to make the comparison fair, we used both full Eigen and Eigen split dataset to make the dataset consistent with other methods. Among all the evaluation measures, the best and the second best ones were produced by our H-Net model, which indicates that our model can learn from the geometry constraints and benefits from the optimal transport solution, achieving state-of-the-art depth predictions. For the quantitative results, we can see that the depth maps generated by our model contained more details, \textit{i.e.} the structural characteristics of buildings, protruding kerbs, bushes, and trees. Besides, our model could effectively distinguish different parts of every object, for example, the upper part of the tree is no longer uniform but is full of outlines and details. 

\subsection{KITTI Ablation Study Results}
\label{ablation_section}
The results of the ablation study on the KITTI dataset are shown in Table \ref{Abalation study}. We can see that the backbone Monodepth2 model \cite{godard2019digging} performed the worst without any of our contributions but by changing the architecture to a Siamese encoder- Siamese decoder, the evaluation measures steadily improved. The reason might be that fusing the complementary information between the stereo image pair gave the framework higher chance to generate accurate predicted depth maps. Our MEA and OT modules were all incorporated in the SE-SD architecture. Row 4 shows that the addition of MEA benefits the depth estimation performance in all the evaluation measures, especially on metrics that are sensitive to large depth errors \textit{e.g.} RMSE. The significantly large improvement of the SE-SD architecture with MEA, is likely due to the epipolar constraint, which allowed the network to learn strong correspondences limited on the same epipolar lines in the rectified stereo images. The impact of the OT-MNL is presented in Row 5, compared to the SE-SD we still can notice a dramatic increase in most of the evaluation metrics. The reason might be that the optimal transport algorithm further improved the MEA by increasing the correct correspondence weights, merging the semantic features while suppressing outliers. In the last row, by combining the backbone with all of our components, the effectiveness of the final framework was significantly improved, as expected, and state-of-the-art results were observed. Besides, although our OT-MEA module was inspired by the MNL, our results outperformed the same SE-SD architecture with MNL. Apart from the performance evaluation measures, we also estimated the number of parameters for each of the examined settings. While all of our proposed components contributed to the overall performance in the self-supervised depth estimation task, the number of parameters was barely increased. We can see from Table \ref{parameters} that our OT-MEA module cost 0.6 million (2.0\%) additional parameters compared with the pure SE-SD architecture.

\subsection{Cityscapes results}
The performance of H-Net has been further evaluated on the Cityscape dataset. 
The results in Figure~\ref{Fig:cityscape} show the accuracy of the depth estimated by H-Net compared to the ground truth, with detailed reconstructions of objects such as cars, human, and trees. More experimental results could be found in the supplementary material.


\section{Conclusion}

In this paper we presented a novel network, the H-Net, for self-supervised depth estimation, achieving state-of-the-art depth prediction. By designing the Siamese encoder-Siamese decoder architecture, exploiting the mutual epipolar attention, and formulating the optimal transport problem, both the global-range correspondence between stereo image pairs and strongly related feature correspondences satisfying the epipolar constraint in the rectified images were effectively explored and fused. We showed how this benefited the overall performance on public datasets and how together they gave a large improvement in evaluation measures, indicating that the model effectively tackled the limits of other self-supervised depth estimation methods and closed the gap with supervised approaches.

{\small
\bibliographystyle{ieee_fullname}
\bibliography{ref}
}

\end{document}